# Dynamic Pricing for On-Demand DNN Inference in the Edge-AI Market

Songyuan Li, Jia Hu, Geyong Min, Haojun Huang, and Jiwei Huang

***Abstract*—The convergence of edge computing and Artificial Intelligence (AI) gives rise to Edge-AI, which enables the deployment of real-time AI applications at the network edge. A key research challenge in Edge-AI is edge inference acceleration, which aims to realize low-latency high-accuracy Deep Neural Network (DNN) inference by offloading partitioned inference tasks from end devices to edge servers. However, existing research has yet to adopt a practical Edge-AI market perspective, which would explore the personalized inference needs of AI users (e.g., inference accuracy, latency, and task complexity), the revenue incentives for AI service providers that offer edge inference services, and multi-stakeholder governance within a market-oriented context. To bridge this gap, we propose an Auction-based Edge Inference Pricing Mechanism (AERIA) for revenue maximization to tackle the multi-dimensional optimization problem of DNN model partition, edge inference pricing, and resource allocation. We develop a multi-exit device-edge synergistic inference scheme for on-demand DNN inference acceleration, and theoretically analyze the auction dynamics amongst the AI service providers, AI users and edge infrastructure provider. Owing to the strategic mechanism design via randomized consensus estimate and cost sharing techniques, the Edge-AI market attains several desirable properties. These include competitiveness in revenue maximization, incentive compatibility, and envy-freeness, which are crucial to maintain the effectiveness, truthfulness, and fairness in auction outcomes. Extensive simulations based on four representative DNN inference workloads demonstrate that AERIA significantly outperforms several state-of-the-art approaches in revenue maximization. This validates the efficacy of AERIA for on-demand DNN inference in the Edge-AI market.**

## I. Introduction

We have been witnessing the unprecedented evolution of Artificial Intelligence (AI) techniques, which drive a variety of cutting-edge applications including Industry 4.0 [1], smart healthcare [2], intelligent transportation [3], etc., that greatly benefit our daily life. The technological breakthrough in Deep Neural Networks (DNNs) enables almost human-level accuracy and automated decision-making [4] for various types of tasks, such as object recognition and robotic control. Edge computing is an emerging distributed computing paradigm that shifts computation power from clouds to the network edge (e.g., edge servers) close to end devices. Consequently, service requests can be processed at the network edge with much lower latency and data transfer overhead, significantly facilitating the realization of real-time, mission-critical services. DNNs are based on complex, multilayer inter-neural topologies, requiring significant computational resources to operate. Using edge computing, some of the heavy computational workload of DNN inference tasks can be offloaded from end devices to edge servers instead of remote clouds, thereby sharply reducing the end-to-end service latency and data transfer costs for AI applications. Thus, the combination of AI and edge computing leads to a burgeoning AI paradigm, known as Edge-AI, which will be revolutionary for many innovative applications like autonomous driving and smart manufacturing. In Edge-AI, one key functionality is device-edge synergistic inference, which aims to mitigate the heavy computation overhead of DNN inference tasks on end devices (e.g., a typical DNN-based object recognition task requires 2.5 Teraflops [5]) through offloading partitioned inference tasks to edge servers. Specifically, the computation-friendly DNN model partitions are locally processed on end devices, while the remaining DNN model partitions with a larger size and heavier computation overhead are offloaded and processed at the edge server.

Edge inference offloading primarily studies how to schedule inference requests and allocate edge resources accordingly, for minimizing the DNN inference latency while meeting the requirements of inference accuracy. Several specialized operations for edge inference acceleration, including DNN model partition [6], [7], [8], [9], DNN model parallelism [10], and early-exiting point selection [11], [12], [13], [14], [15], etc., have been studied in the literature. Nevertheless, the existing research has yet to study the problem from the Edge-AI market perspective, which would comprehensively consider the personalized inference requirements of AI users (e.g., inference accuracy, latency, and task complexity), the revenue incentives for AI service providers offering edge inference services, and multi-stakeholder governance in a market-oriented context.

The Edge-AI market presents a more realistic and sophisticated service scenario that needs to be comprehensively explored in these aspects:

- Different AI users make DNN inference requests with diversified task complexity and personalized performance requirements, in terms of DNN inference accuracy and latency. Hence, it is imperative to have an effective edge resource management strategy to cope with a large number

*Corresponding author: Jia Hu and Geyong Min.*

Songyuan Li, Jia Hu, and Geyong Min are with the Department of Computer Science, Faculty of Environment, Science and Economy, University of Exeter, Exeter EX4 4PY, U.K. (e-mail:{S.Y.Li, J.Hu, G.Min}@exeter.ac.uk).

Haojun Huang is with the School of Electronic Information and Communications, Huazhong University of Science and Technology, Wuhan 430074, China (e-mail: hjhuang@hust.edu.cn).

Jiwei Huang is with the Beijing Key Laboratory of Petroleum Data Mining, China University of Petroleum, Beijing 102249, China (e-mail: huangjw@cup.edu.cn).

of diverse inference requests.

- The Edge-AI market involves multiple stakeholders, including the AI service providers, AI users, and edge infrastructure provider. The AI service providers expect to maximize their revenue at a certain rental cost of edge infrastructures. However, the AI users aim to benefit from high-quality edge inference services while minimizing budget expenditure. To coordinate these conflicting interests, it necessitates a truthful and fair edge inference trading environment that can regulate the market behaviors of different stakeholders.

To tackle these research challenges, we adopt the multi-exit DNN inference to meet the personalized inference requirements of different AI users. Specifically, low-complexity inference tasks are enabled to exit early at the shallow DNN layer upon correct classification. This significantly reduces the inference latency and required edge inference resources for such tasks, while freeing up finite edge computation capacity to accelerate the inference process of high-complexity tasks and accommodate more inference tasks. Furthermore, in the Edge-AI market, the AI service providers play as the auctioneer to operate our proposed Auction-based Edge Inference Pricing Mechanism namely AERIA, making multi-dimensional decisions on DNN model partition, edge inference pricing, and resource allocation. Edge inference demand analysis is rigorously conducted to derive the users' edge resource requests that align with their personalized inference requirements, enabling on-demand edge inference resource allocation. The bid-winning AI users acquire demanded edge inference resources at the minimum winning price, while the AI service providers increases the revenue from competitive auction. Thus, a truthful and fair edge inference trading environment is created in the Edge-AI market, which is theoretically secured by the auction properties of incentive compatibility (i.e., *preventing the market manipulation*) and envy freeness (i.e., *preserving fair trade practice*). In a nutshell, this article makes the following contributions:

- To the best of our knowledge, we are the first to perform comprehensive system modeling of the Edge-AI market, which characterizes the market interactions amongst the AI service providers, AI users, and edge infrastructure provider. Empowered by our multi-exit device-edge synergistic inference design, the AI service providers offer personalized edge inference services, addressing diversified inference requirements of AI users, in terms of DNN inference accuracy, latency, and task complexity.
- We develop a novel Auction-based Edge Inference Pricing Mechanism (AERIA) for revenue maximization, to handle the multi-dimensional optimization problem on DNN model partition, edge inference pricing and resource allocation. Our AERIA mechanism is theoretically guaranteed to be incentive compatible and envy-free, while approximately reaching revenue maximization in a good competitive ratio. The strategic AERIA mechanism design ensures the effectiveness, truthfulness, and fairness of edge inference trading in the Edge-AI market.
- Our proposed AERIA mechanism is empirically evaluated through extensive simulation experiments based on real-world datasets. Compared with several state-of-the-art approaches, the experimental results demonstrate that our AERIA mechanism outperforms in revenue maximization by approximately 60%, while providing high-quality edge inference services. The code of AERIA is available at: https://github.com/songyuanli/AERIA.

The rest of this article is organized as follows. Section II introduces the related work. Section III presents the system model. Section IV analyzes the multi-exit synergistic inference latency and formulates the multi-dimensional optimization problem. Section V describes our proposed AERIA mechanism in details. Section VI discusses the experimental results. Section VII concludes this article.

## II. Related Work

**Synergistic DNN Inference:** It focuses on the reduction of DNN inference latency by streamlining inference tasks and integrating heterogeneous computational resources. Chen *et al.* [6] proposed a joint DNN model partition and inference offloading approach based on swarm optimization, which balanced the tradeoff between energy consumption and inference latency in collaborative IoT-edge-cloud environments. Xu *et al.* [7] utilized the deep reinforcement learning technique, and then designed an online inference offloading algorithm for multiple requests in mobile edge clouds, with the objective of admitting the maximum edge inference requests. Li *et al.* [9] dynamically partitioned and streamlined the DNN inference tasks, thereby achieving throughput-optimized DNN inference acceleration in mobile edge computing. Tang *et al.* [8] studied the multi-user DNN partitioning and inference resource allocation issue in collaborative device-edge environments, in which the minimax optimization was applied to optimize the multi-user inference latency. Exploring the potential of edge-cloud collaboration, Zhuang *et al.* [10] proposed a delay-aware DNN partition algorithm that leveraged pipeline-based parallel DNN execution, thereby enabling efficient edge-cloud inference acceleration.

**Early-Exit DNNs:** It inherently accelerates the DNN inference process by optimizing the DNN architecture itself. Specifically speaking, BranchyNet [16] initially proposed a modified DNN architecture that enabled early exits from the DNN network via branches as soon as a certain confidence criterion was reached. Li *et al.* [11] applied the early-exit DNNs for rightsizing the vanilla DNN, thereby reducing the inference latency via exiting at an intermediate DNN layer. Extending the prototypical design of early-exit DNNs, Henna *et al.* [12] developed a branchy architecture for graphical neural networks (GNNs) to accelerate the distributed GNN training and inference process at edge. Dong *et al.* [13] put forward a multi-exit DNN inference acceleration scheme with multi-dimensional optimization on early-exit selection, DNN model partition and edge resource allocation, where the correlation between early-exit settings and synergistic inference was sufficiently investigated. Liu *et al.* [14] leveraged the early-exit DNN to adaptively hasten the edge offloading of DNN inference task streams, in accordance with volatile edge computing environments.

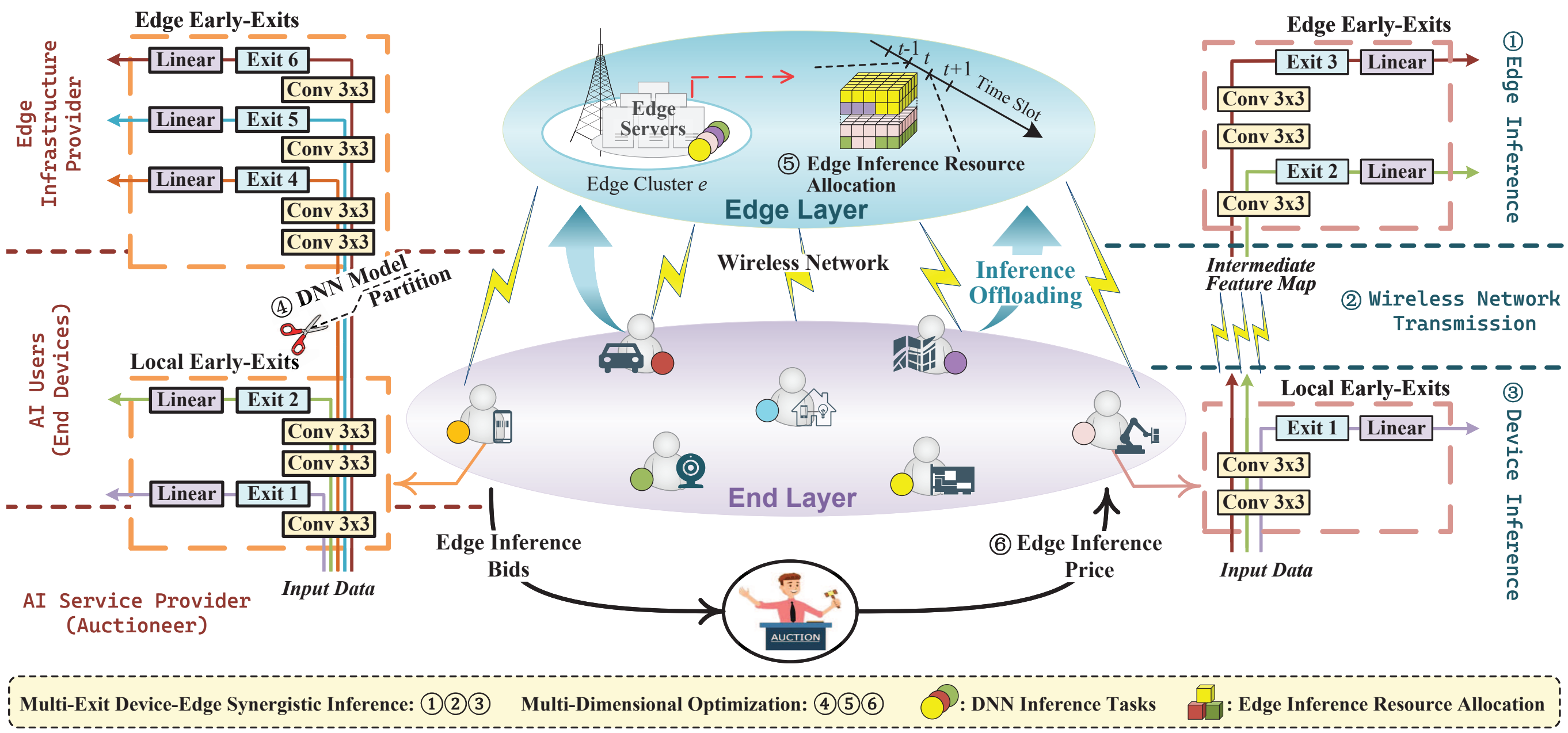

Fig. 1: Multi-exit DNN Inference and Auction Dynamics in the Edge-AI Market.

**Dynamic Pricing for Edge Resources:** It has become a fundamental issue in mobile edge computing, considering the economic incentives of end users and service providers. Wang *et al.* [17] proposed a dynamic pricing approach based on mean field games for edge computing, which jointly optimized the task scheduling cost and the edge operator's welfare. Ma *et al.* [18] designed a combinatorial double auction mechanism in device-edge environments, with a Vickrey Clarke Groves (VCG)-based pricing strategy developed to ensure the incentive compatibility. Wang *et al.* [19] designed a Stackelberg game-based dynamic service pricing mechanism in edge computing, which optimized the service utility of edge servers and the users's Quality of Service (QoS) gains. More recently, reinforcement learning has been leveraged for dynamic pricing decisions in edge resources. Tang et al. [20] developed a container scheduling mechanism for edge computing, using deep reinforcement learning to achieve continuous profit maximization for service providers. Tutuncuoglu et al. [21] employed a reinforcement learningbased function approximator to adaptively determine revenue-optimal pricing for edge computing resources under varying user workloads. He et al. [22] designed a reinforcement learning-based pricing module for edge resources to incentivize operators in supporting low-latency video analytics at the network edge. Li et al. [23] applied reinforcement learning to bidding dynamics in edge computing, deriving a welfare-optimal multi-user resource allocation solution.

In this work, we adopt a multi-exit synergistic device-edge inference framework in the Edge-AI market, and propose an auction-based edge inference pricing mechanism for revenue maximization. Differing from the literature, our auction mechanism takes account of the specialized operations for edge inference acceleration, including DNN model partition and multi-exit inference. Furthermore, we comprehensively addresses various service trading properties (revenue efficiency, truthfulness, and fairness) from an ecosystem perspective. Although reinforcement learning-based pricing approaches offer strong adaptability to dynamic edge resource bidding environments, they often operate as "black-box" models, making it difficult for decision-makers to interpret or justify their pricing strategies. Such lack of transparency can further undermine truthfulness and fairness in edge resource trading. To the best of our knowledge, this research addresses a technical void in the Edge-AI market. It thoroughly considers several significant factors in the Edge-AI market, including the personalized inference requirement of AI users, the revenue incentive for the AI service providers, as well as the truthfulness and fairness of edge inference trading, while performing multi-dimensional system optimization.

## III. System Model

### A. The Overview of Edge-AI Market

**System Outline:** As illustrated in Fig. 1, we envision an Edge-AI market with the interactions among AI service providers, AI users, and the edge infrastructure provider. Multiple AI service providers collaboratively share edge infrastructures to offer different Edge-AI services, implemented by various DNN models for distinct inference tasks. The Edge-AI market operates at a time-slotted manner, with the time horizon divided into discrete time slots of length $\tau$, indexed by $t$. At the beginning of each time slot $t$, each AI user determines whether to place an edge inference bid to the AI service providers, detailing the requested Edge-AI service along with the performance requirements for inference latency and accuracy. Multiple AI service providers jointly act as an auctioneer to determine the bid-winning AI users who are granted access to edge inference services. Based on the submitted bids, they dynamically price the edge inference service as $p_t$ with the fluctuation of supply and demand for edge inference resources. The rental cost of edge infrastructures is taken into account when setting the edge inference price $p_t$, as edge inference resources are leased from the edge infrastructure provider.

**Multi-Exit Synergistic Device-Edge Inference:** As shown in Fig. 1 ①②③, the AI users at the end layer have their inference requests processed at an accelerated rate, via subscribing the high-performance inference services deployed at the edge layer. Following the synergistic device-edge inference scheme, the AI service providers mainly handle the hard-to-infer part (i.e., deep neural layers) by hiring computational resources from the edge infrastructure provider. Meanwhile, the AI users primarily handle the simple-to-infer part (i.e., shallow neural layers) at their local end devices. To meet the personalized requirements of DNN inference, the multi-exit DNNs (ME-DNNs) are deployed on end devices and edge servers. With multiple classifier heads (e.g., `nn.Linear`) as exit branches, ME-DNNs enable on-demand inference services for varying task complexity. The low-complexity input data samples would be correctly classified at local exits (i.e., exit points on end devices), without the need to transmit inference requests to the edge. Complex inference tasks are offloaded to the edge but can finish early at edge exits with an acceptable accuracy, which reduces the inference latency and eliminates the redundant computations in deeper neural layers.

**Multi-Dimensional Optimization:** As shown in Fig. 1 ④⑤⑥, the AI service providers optimize the multi-exit synergistic device-edge inference process via multi-dimensional decisions on DNN model partition, edge inference pricing and resource allocation. The joint solution for DNN model partition and edge inference resource allocation is customized to each AI user, which satisfies the user's personalized performance requirement on inference latency and accuracy. As an auctioneer, the AI service providers dynamically price the edge inference services for revenue maximization, based on the market competence demonstrated by AI users in their edge inference bids. The dynamic pricing process follows the principle of market economics [24], [25], depending on the demand and capacity ratio of edge inference resources. The bid-winning AI user will enforce a customized DNN model partition solution, accordingly allocated edge inference resources commensurate with the user's edge resource demand.

### *B. System Formulation*

**AI Service Providers (Auctioneer):** Different AI service providers offer $D$ types of ME-DNNs, which are preloaded onto edge servers and end devices, denoted by $\mathcal{M} = \{M_1, ..., M_j, ..., M_D\}$. Before inference deployment, these ME-DNNs are trained on large-scale datasets ready for inference. During inference runtime, a flexible DNN model partitioning approach is employed, where the DNN portions for device/edge inference can be extracted directly from the preloaded ME-DNN. As exemplified in Fig. 1, an ME-DNN model is composed of one main branch and multiple exit branches. To accelerate the DNN inference via early exits, different exit branches are adhered to the main branch at distinct DNN layers. Note that, the non-cascade concatenated neural layers, such as, residual blocks [26] and inception blocks [27], are integrated as a layer unit, where the exit branches are inserted between these layer units. In general, an ME-DNN model is defined by $M_j = (L_j, B_j)$, where

- $L_j = \{l_{j,1}, l_{j,k}, ..., l_{j,g_j}\}$ is the set of DNN layers constituting the main branch, in which $g_j$ denotes the number of DNN layers. A DNN layer $l_{j,k}$ is defined as $\langle O_{j,k}, \rho_{j,k}\rangle$, where $O_{j,k}$ is the output data size of DNN layer $l_{i,k}$, and $\rho_{j,k}$ denotes the computation overhead of DNN layer $l_{i,k}$.
- $B_j = \{b_{j,k} | k \in \Gamma_j\}$ is the set of exit branches, in which $\Gamma_j$ records the embedding position of different exit branches along the main branch. An exit branch $b_{j,k}$ is formulated as $\langle \nu_{j,k}(\sigma), \varrho_{j,k}\rangle$, where $\nu_{j,k}(\sigma)$ indicates the expected exiting probability of exit branch $b_{j,k}$ with respect to an inference confidence criterion $\sigma$ [28], and $\varrho_{j,k}$ specifies the computation overhead of exit branch $b_{j,k}$. Note that $\sum_{k\in\Gamma_j} \nu_{j,k}(\sigma) = 1$.

The computation overhead of DNN layers and exit branches is quantified by the floating point of operations. Meanwhile, the inference confidence criterion $\sigma$ is defined by a softmax cross-entropy threshold of exit outputs [28], which characterizes the inference accuracy requirement of AI users. Intuitively, $\sigma$ reflects the AI user's desired level of prediction correctness. If the softmax cross-entropy output of the classifier head at a given exit branch meets the threshold $\sigma$, the DNN inference process terminates at that exit; otherwise, it continues to a deeper main branch layer. A stringent inference confidence criterion $\sigma$ generally implies a higher inference accuracy requirement for AI users. The exit branches on the end device are called *local exits*, while those on the edge server are referred to as *edge exits*.

The expected exiting probability $\nu_j^k(\sigma)$ is acquired through offline inference experiments from the validation set [13]. With all validation samples inferred from input to all exits by the confidence criterion $\sigma$, the number of early-exit samples at each exit branch $b_{j,k}$ is counted as $x_{j,k}(\sigma)$. The expected exiting probability $\nu_j^k(\sigma)$ is obtained as $x_{j,k}(\sigma)/X$, where $X$ is the total number of validation samples. The obtained exiting probability in expectation tends to be precise with sufficient validation samples processed. Typically, a stringent inference confidence criterion $\sigma$ reduces the exiting probability $\nu_j^k(\sigma)$ at shadow exits, pushing deeper DNN inference. By contrast, a relaxed $\sigma$ increases the expected likelihood of exiting at earlier DNN layers.

In the Edge-AI market, the AI service providers share leased edge infrastructures to offer users a range of edge inference services driven by various ME-DNNs. As an *auctioneer*, the AI service providers periodically collect the edge inference bids from various AI users across time slots, and determine the bid-winning AI users to be served at the edge. According to the latest collected bidding information, the multi-dimensional decisions on DNN model partition, edge inference pricing and resource allocation are dynamically made in an online fashion at each time slot $t$ for revenue maximization.

**AI Users/End Devices[1] (Bidders):** There are totally $N$ AI users/end devices who place the edge inference bid across time slots, denoted by $\mathcal{U} = \{u_1, .., u_i, ..., u_N\}$. At each time slot $t$, there are $N_t$ AI users (i.e., $\mathcal{U}_t \subset \mathcal{U}$) who bid for edge inference services. Then, a subset of $\widehat{N}_t < N_t$ AI users, represented by $\widehat{\mathcal{U}}_t \subset \mathcal{U}_t$, will eventually win the auction and pay the AI service providers for edge inference services. The remaining AI users $\mathcal{U}_t/\widehat{\mathcal{U}}_t$ resort to locally processing their inference computation on their devices. With no loss of generality, each AI user is assumed to simply place an edge inference bid within a time slot, and the AI users who concurrently propose multiple edge inference bids can be regarded as a band of AI users. The computation capacity $f_i$ of end device $u_i$ is measured by the floating point operations per second (FLOPS).

[1] In this article, we speak interchangeably of the AI user and the end device.

TABLE I: Summary of Key Notions

| Symbol | Definition |
|---|---|
| $\tau, t$ | Duration, index of time slot. |
| $\mathcal{M}, D, M_j$ | Set, number, index of ME-DNN models. |
| $\rho_{j,k}, \mu_{j,k}$ | Computation overhead, forward-propagation probability of DNN layer $l_{j,k}$. |
| $\varrho_{j,k}, \nu_{j,k}$ | Computation overhead, exiting probability of exit branch $b_{j,k}$. |
| $\mathcal{U}, N, u_i$ | Set, number, index of AI users (end devices). |
| $\mathcal{U}_t$ | Set of AI users who bid for edge inference services at the time slot $t$. |
| $f_i, f_e$ | Computation capacity of end device $u_i$ and edge cluster $e$, measured by FLOPS. |
| $R_i = \{\beta_i, t_i, \sigma_i, w_i\}$ | Edge inference bid proposed by the AI user $u_i$, which specifies the bidding budget $\beta_i$, the performance requirement on inference latency $t_i$ and accuracy $\sigma_i$, and the requested ME-DNN model $M_{w_i}$. |
| $p_t^r, p_t^e$ | Edge infrastructure's rental price, and electricity price at the time slot $t$. |
| $F_i^{dev}, F_i^{edge}$ | Computation overhead of the AI user $u_i$'s inference request occurring during the phases of device inference and edge inference. |
| $T_i, T_i^{\text{dev}}, T_i^{\text{edge}}, T_i^{\text{net}}$ | Overall synergistic inference latency, device-inference latency, edge-inference latency, network transmission latency of the AI user $u_i$. |
| $\widetilde{s}_{i,t}$ | DNN model partition request of the AI user $u_i$, associated with $\widetilde{a}_{i,t}$. |
| $\widetilde{a}_{i,t}$ | Edge inference resource request of the AI user $u_i$ at the time slot $t$. |
| $s_{i,t}$ | DNN model partition decision for the AI user $u_i$ at the time slot $t$. |
| $a_{i,t}$ | Edge inference resource allocation for the AI user $u_i$ at the time slot $t$. |
| $p_t$ | Edge inference price per unit FLOPS at the time slot (billing cycle) $t$. |

At each time slot $t$, the edge inference bid of AI user $u_i \in \mathcal{U}_t$ is formulated as $R_i = \{\beta_i, t_i, \sigma_i, w_i\}$, where the bidding budget $\beta_i$, the performance requirement on inference latency $t_i$ and accuracy $\sigma_i$, as well as the requested ME-DNN model type $M_{w_i} \in \mathcal{M}$ are specified. Specifically, we extend the previously introduced notion of inference confidence criteria $\sigma$ to $\sigma_i$, capturing heterogeneous inference accuracy requirements of different AI users $u_i$. By means of multi-exit device-edge synergistic inference, the inference confidence gained for user $u_i$ should be no worse than $\sigma_i$, while the corresponding inference latency is no more than $t_i$. Otherwise, the AI user $u_i$ will decline the edge inference services.

The inference tasks a user $u_i$ would run on its own devices depend on various factors, encompassing the user's bidding budget $\beta_i$ for edge inference services, the local computation capacity $f_i$, as well as the performance requirements on inference latency $t_i$ and accuracy $\sigma_i$. Therefore, the AI users with stringent performance requirements are inclined to offload more inference computation tasks to the edge even their devices may possess better computational capacity than others, provided they have sufficient bidding budget.

**Edge Infrastructure Provider:** Edge computation resources are operated as an edge cluster $e$ comprising a set of co-located edge servers that could be mixed GPU/CPU compute nodes, and are hired by the AI service providers for edge inference. The computation capacity of edge cluster $e$ is $f_e$ in FLOPS, and the wireless communication distance from the AI user (end device) $u_i$ is $\triangle_{i,e}$. A whole set of ME-DNN models $\mathcal{M}$ is proactively deployed at the edge cluster $e$. In the edge cluster $e$, inter-server resource coordination is performed internally [29][30], enabling inference task workloads to be balanced across edge servers and enhancing the entire cluster's resource utilization.

To demonstrate the generality and applicability of our proposed approach, we take the deregulated energy market [31] into account, where the electricity price $p_t^e$ is stochastically evolved in accordance with the time-varied energy demand and supply. The operational cost of edge infrastructure provider primarily derives from the energy cost of edge servers [32][33]. Accordingly, the edge infrastructure provider determines the edge infrastructure's rental price $p_t^r$ at each time slot $t$, which can fluctuate in response to volatile energy costs in the deregulated market. Alternatively, the fixed-pricing policy adopted in regulated energy markets can be also applied into our approach.

## IV. Problem Formulation

### A. Multi-Exit Synergistic Inference Latency Analysis

The multi-exit synergistic inference latency for each AI user $u_i$ is associated with the joint decision on DNN model parition $s_{i,t} = \{0, 1, ..., g_{w_i}\}$, and edge inference resource allocation $a_{i,t} \in [0, f_e]$. According to the device-edge inference framework in Fig. 1, the overall synergistic inference latency derives from three phases, which are 1) device inference, 2) wireless network transmission, and 3) edge inference.

**Computation Overhead of Multi-exit DNN Inference:** Intuitively, the multi-exit inference latency is positively correlated to the computation overhead of DNN layers and exit branches. Owing to the non-deterministic exit selection during ME-DNN inference, the redundant computation overhead of multi-exit inference is reflected by the exiting probability $\nu_{j,k}(\sigma_i)$. In other words, the exiting probability $\nu_{j,k}(\sigma_i)$ suggests the probabilistic forward propagation of ME-DNN. For the ME-DNN model $M_j$, the forward-propagation probability $\mu_{j,k}$ through its $k$-th DNN layer (i.e., $l_{j,k}$) can be formulated as ($\Gamma_j^{[k_1,k_2]} = \{x \in \Gamma_j | k_1 \le x \le k_2\}$):

$$\mu_{j,k} = 1 - \sum\nolimits_{x \in \Gamma_j^{[1,k-1]}} \nu_{j,x}(\sigma_i). \tag{1}$$

The DNN model partition decision $s_{i,t}$ suggests that the end device $u_i$ infers the 1-st to $s_{i,t}$-th DNN layers and their attached exit branches, while the rest of DNN layers and exit branches are offload for edge inference. Meanwhile, the ME-DNN computation overhead arises from two parts, including main-branch inference and exit-branch inference. Correspondingly, the expected computation overhead of device inference after DNN model partitioning $s_{i,t}$ is formulated as ($j = w_i$):

$$F_i^{\text{dev}}(s_{i,t}) = \overbrace{\sum\nolimits_{k=1}^{s_{i,t}} (\mu_{j,k} \cdot \rho_{j,k})}^{\text{Main-branch Inference}} + \overbrace{\sum\nolimits_{k \in \Gamma_j^{[1,s_{i,t}]}} (\nu_{j,k} \cdot \varrho_{j,k})}^{\text{Local-exit Inference}}. \tag{2}$$

Likewise, the expected computation overhead of edge inference is expressed ($j = w_i$):

$$F_i^{\text{edge}}(s_{i,t}) = \overbrace{\sum\nolimits_{k=s_{i,t}+1}^{g_j} (\mu_{j,k} \cdot \rho_{j,k})}^{\text{Main-branch Inference}} + \overbrace{\sum_{k \in \Gamma_j^{[s_{i,t}+1, g_j]}} (\nu_{j,k} \cdot \varrho_{j,k})}^{\text{Edge-exit Inference}}. \tag{3}$$

**Device Inference Latency:** The shallow DNN layers and their attached exit branches are locally processed. Combined with the expected computation overhead in Eq. (2), the shallow-layer inference execution latency $T_i^{\text{dev}}$ at end devices is derived as:

$$T_i^{\text{dev}}(s_{i,t}) = F_i^{\text{dev}}(s_{i,t}) \ / \ f_i. \tag{4}$$

**Network Transmission Latency:** Suppose that the DNN inference task does not early exit at the end device, the intermediate feature map generated by the $s_{i,t}$-th DNN layer will be transmitted to the edge cluster $e$ for deep inference. Let $\xi_i$ denote the average propagation delay, and $r_{i,t}$ the time-varying wireless data rate between edge cluster $e$ and device $u_i$ over time slots $t$. Considering the forward-propagation probability $\mu_{j,s_{i,t}}$, and the feature-map size $O_{j,s_{i,t}}$ at the partitioned DNN layer $l_{j,s_{i,t}}$, the network transmission latency can be formulated as ($j = w_i$):

$$T_i^{\text{net}}(s_{i,t}) = \mu_{j,s_{i,t}} \cdot \left( \xi_i + \frac{O_{j,s_{i,t}}}{r_{i,t}} \right). \tag{5}$$

where $r_{i,t} = 0$ represents intermittent network interruptions resulting in $T_i^{\text{net}} = \infty$.

**Edge Inference Latency:** The edge cluster $e$ receives the intermediate feature map from the end device $u_i$, after which the edge computation resources are allocated for AI users for deep inference. Given the allocated amount of edge inference resource $a_{i,t} \in (0, f_e]$ for each bid-winning AI user $u_i \in \widehat{\mathcal{U}}_t$, the deep inference execution latency at edge is expressed as:

$$T_i^{\text{edge}}(s_{i,t}, a_{i,t}) = F_i^{\text{edge}}(s_{i,t}) \ / \ a_{i,t}. \tag{6}$$

To summarize, the overall latency of multi-exit synergistic device-edge inference, which is composed of the above three phases, can be expressed as:

$$T_i(s_{i,t}, a_{i,t}) = T_i^{\text{dev}}(s_{i,t}) + T_i^{\text{net}}(s_{i,t}) + T_i^{\text{edge}}(s_{i,t}, a_{i,t}). \tag{7}$$

### B. Multi-Dimensional Optimization Problem

Considering the multi-user competition for limited edge inference resources, we address the joint edge inference scheduling and pricing problem with multiple decision-making dimensions, including 1) edge inference pricing, 2) DNN model partition decision, and 3) edge inference resource allocation. The multi-dimensional optimization problem targets revenue maximization, from the standpoint of AI service providers who provide edge inference services. According to the user's personalized performance requirements on inference accuracy and latency across time slots, each AI user $u_i \in \mathcal{U}_t$ periodically figures out its latest DNN model partition request $\widetilde{s}_{i,t}$ and the edge inference resource request $\widetilde{a}_{i,t}$. Responding to the demands for edge inference resources $\{\widetilde{a}_{i,t} | u_i \in \mathcal{U}_t\}$ to serve the DNN fragments (as specified by $\{\widetilde{s}_{i,t} | u_i \in \mathcal{U}_t\}$), the AI service providers jointly act as an auctioneer who determines if approving the AI user $u_i$'s DNN model partition request $\widetilde{s}_{i,t}$ and edge inference resource request $\widetilde{a}_{i,t}$, and dynamically sets the edge inference price $p_t$ for revenue maximization. The bid-winning AI users $u_i \in \widehat{\mathcal{U}}_t$ ($\widehat{\mathcal{U}}_t \subset \mathcal{U}_t$) with sufficient budget $\beta_i$ will be approved to be served at the edge, while the other AI users $u_i \in \mathcal{U}_t \backslash \widehat{\mathcal{U}}_t$ would be dismissed, as formulated by this equation:

$$\begin{cases} a_{i,t} = \widetilde{a}_{i,t}, s_{i,t} = \widetilde{s}_{i,t}, \forall\, u_i \in \widehat{\mathcal{U}}_t \ \text{ whose } \ p_t \cdot \widetilde{a}_{i,t} \le \beta_i \\ a_{i,t} = 0, s_{i,t} = \text{NULL}, \forall\, u_i \in \mathcal{U}_t \backslash \widehat{\mathcal{U}}_t \ \text{ whose } \ p_t \cdot \widetilde{a}_{i,t} > \beta_i \end{cases} \tag{8}$$

where $s_{i,t}$ and $a_{i,t}$ respectively represent the DNN model partition decision and the edge inference resource allocation for the AI user $u_i$ at the time slot $t$, as shown in Table I.

Let $\{p_t, \mathbf{S}_t, \mathbf{A}_t\}$ represent the optimal decision vector for all bidding AI users $\mathcal{U}_t$ at the time slot $t$, in which $\mathbf{S}_t = \{s_{i,t} | u_i \in \mathcal{U}_t\}$ and $\mathbf{A}_t = \{a_{i,t} | u_i \in \mathcal{U}_t\}$. Then, the cooperative edge inference scheduling and pricing problem at each time slot $t$ is defined as below. Note that $I_{\{condition\}}$ is an indicator function returning 1 if the *condition* is true, otherwise 0.

(**P1**) :

$$\{p_t, \mathbf{S}_t, \mathbf{A}_t\} = \underset{p_t,\, s_{i,t},\, a_{i,t}}{\arg\max} \overbrace{p_t \cdot \sum_{u_i \in \mathcal{U}_t} I_{\{T_i(s_{i,t}, a_{i,t}) \leq t_i\}} \cdot a_{i,t}}^{\text{Revenue } \Pi(\mathcal{U}_t)} \tag{9}$$

$$\text{s.t.} \quad p_t \cdot a_{i,t} \leq \beta_i, \ \forall u_i \in \mathcal{U}_t \tag{9a}$$

$$\sum_{u_i \in \mathcal{U}_t} a_{i,t} \leq f_e \tag{9b}$$

$$p_t \cdot \sum_{u_i \in \mathcal{U}_t} a_{i,t} \geq p_t^r \cdot (1+\gamma) \tag{9c}$$

$$a_{i,t} \in [0, f_e], \ s_{i,t} \in \{0, 1, ..., g_{w_i}\}, \ \forall u_i \in \mathcal{U}_t \tag{9d}$$

The objective function (9) maximizes the revenue $\Pi(\mathcal{U}_t)$ across all time slots. The indicator function $I_{\{T_i(s_{i,t},a_{i,t}) \leq t_i\}}$ denotes that only the bid-winning AI user $u_i$, whose inference latency requirement $t_i$ is satisfied as $T_i(s_{i,t}, a_{i,t}) \leq t_i$, would be charged for edge inference services. Constraint (9a) indicates each AI user $u_i \in \mathcal{U}_t$'s payment for edge inference services is budget-balanced. Constraint (9b) implies the maximum edge inference resource capacity. Constraint (9c) enforces the minimum acceptable profit-rate threshold required by AI service providers, denoted by $\gamma$, while balancing the rental cost of edge infrastructures. Constraint (9d) defines the feasible range of DNN model partition decision $s_{i,t}$ and edge inference resource allocation $a_{i,t}$.

Solving the multi-dimensional problem (**P1**) suffers from significant computational intractability in each decision-making dimension, detailed as follows.

- **DNN Model Partition Decision $\mathbf{S}_t$:** Each AI user $u_i \in \mathcal{U}_t$ selects the optimal DNN model partition decision $s_{i,t}$ from $(g_{w_i}+1)$ candidate partition points, i.e., $\{0, 1, ..., g_{w_i}\}$. Therefore, for $N_t$ AI users, the decision space expands up to $\mathcal{O}(\prod_{u_i \in \mathcal{U}_t}(g_{w_i}+1))$.
- **Edge Inference Resource Allocation $\mathbf{A}_t$:** Suppose that the minimum edge resource allocation interval is $\varphi$, which generates $f_e/\varphi$ edge resource slots. Hence, the decision on edge inference resource allocation $a_{i,t}$ maps to the multiple-choice knapsack problem, which is NP hard with the solution space of $\mathcal{O}(\frac{(\frac{1}{\varphi}+N_t-1)!}{(N_t-1)!\cdot(\frac{f_e}{\varphi})!})$.
- **Edge Inference Pricing $p_t$:** It is strongly coupled with the other two decision dimensions. The decision on $p_t$ regulates each bidding AI user $u_i \in \mathcal{U}_t$'s edge inference demand, i.e., whether to approve the user's DNN model partition request $\widetilde{s}_{i,t}$ and edge inference resource request $\widetilde{a}_{i,t}$ (see Eq. (8)). In return, the finalized decision on $a_{i,t}$ and $s_{i,t}$ jointly reflects the AI user's affordability to edge inference resources under $p_t$, which ultimately contributes to the objective of revenue maximization $\max \Pi(\mathcal{U}_t)$.

### C. Desirable Properties for the Edge-AI Market

When solving the multi-dimensional optimization problem **P1**, we also strive to ensure an effective, truthful and fair edge inference trading environment. In terms of *trading effectiveness*, the AI service providers are guaranteed to earn the approximately maximized revenue from AI users, considering the significant computational challenges of **P1**. Being a step further, each bidding AI user $u_i$ is regulated to *truthfully* report the market profile (e.g., bidding budget $\beta_i$). On-demand edge inference services for each bid-winning AI user is guaranteed to be *fairly* rewarded by the user's charged payment. To achieve these ambitious visions for Edge-AI market, we pursue the following three desirable properties.

- **Competitiveness in Revenue Maximization:** The auction-based edge inference pricing mechanism should derive a feasible solution which approximately maximizes the revenue in a good competitive ratio. The AI service providers can thereby gain the economic incentive to maintain and develop the DNN inference services with better performance.
- **Incentive Compatibility:** No matter what the bidding strategies adopted by the other users, the AI user $u_i$ would regard reporting the true bidding budget $b_i$ as the most advantageous measure to earn the edge inference auction. This property serves as the validity foundation for our auction-based edge inference pricing mechanism, which prevents the market manipulation behaviors and maintains order in the Edge-AI market.
- **Envy Freeness:** Each AI user always prefers its own experienced edge inference service to that of others, specifically in terms of the multi-dimensional decision on DNN model partition $s_{i,t}$, edge inference pricing $p_t$ and resource allocation $a_{i,t}$. This property provides equal treatment amongst bidders (i.e., AI users) and generates a fair Edge-AI market.

## V. Auction Mechanism Design

### A. AERIA Mechanism Overview

The auction-based edge inference pricing mechanism namely AERIA is designed not only to solve the multi-dimensional optimization problem **P1**, but also to strategically satisfy the desirable properties of the Edge-AI market (effectiveness, truthfulness, and fairness). The **P1** objective of *revenue maximization* can be achieved in a good competitiveness ratio, while each AI user $u_i$ who reports the *true bidding budget* $\beta_i$ is theoretically guaranteed to obtain the *envy-free* edge inference resource price and allocation.

As illustrated in Fig. 2, the AERIA mechanism solves the multi-dimensional optimization problem **P1** by uniquely decoupling it into two phases. In Phase I, each AI user $u_i \in \mathcal{U}_t$ firstly conducts edge inference demand analysis to work out its DNN model partition request $\widetilde{s}_{i,t}$ and the associated edge inference resource request $\widetilde{a}_{i,t}$ by solving the optimization subproblem **P2**. In Phase II, the AI service providers utilize the developed Consensus Estimate-based Competitive (COCO) auction algorithm to determine if approving the AI user $u_i$'s DNN model partition request $\widetilde{s}_{i,t}$ and edge inference resource allocation $\widetilde{a}_{i,t}$ and to set the edge inference price $p_t$, therefore finalizing the multi-dimensional decision on $\{p_t, \mathbf{S}_t, \mathbf{A}_t\}$ for problem **P1**. The AERIA mechanism is operated at each time slot $t$ in an online manner, according to the latest edge inference bids from AI users. Specifically,

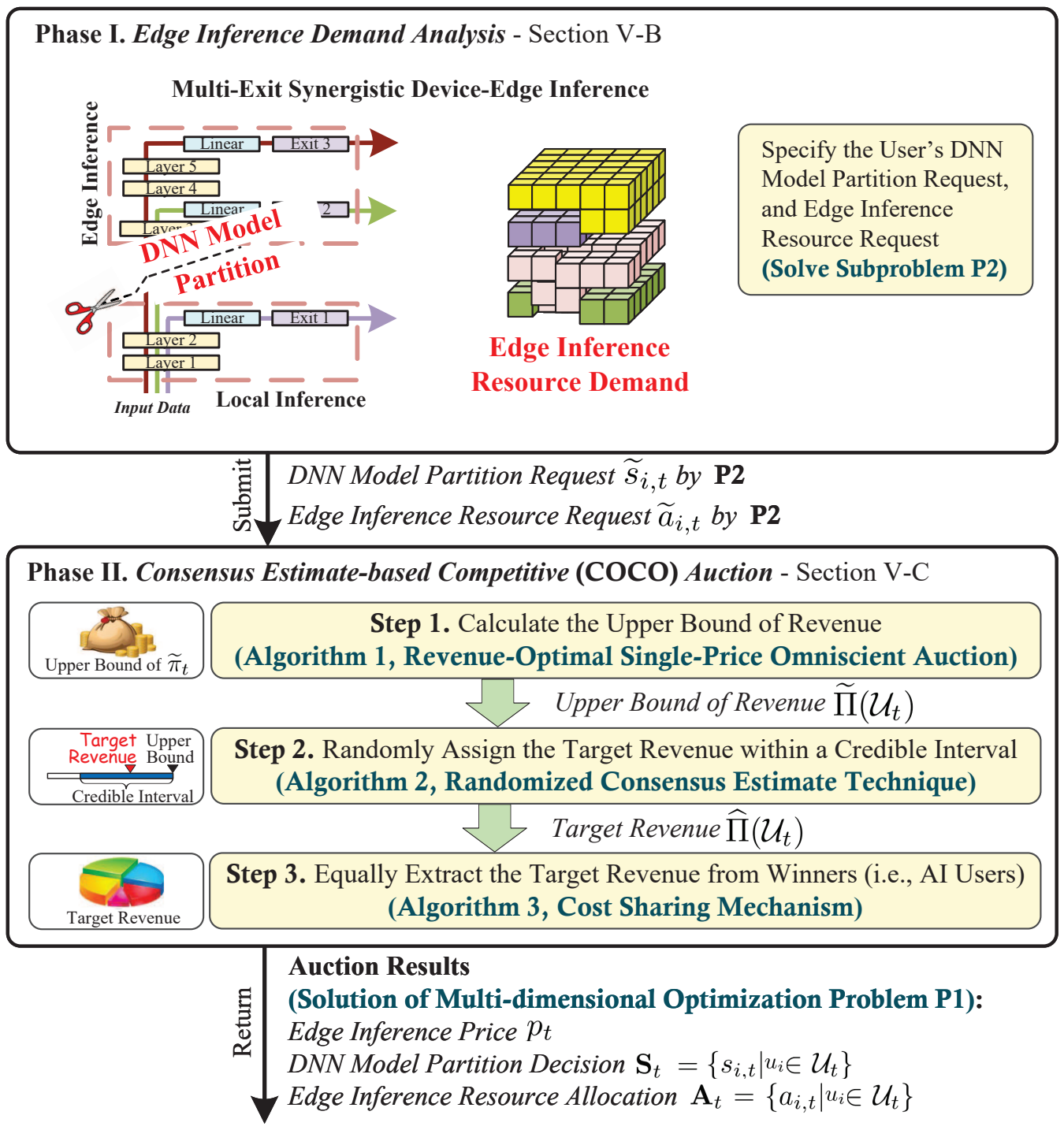

Fig. 2: Overview of Auction-based Edge Inference Pricing Mechanism (AERIA).

- **Phase I (Edge Inference Demand Analysis):** Each AI user $u_i$ expects to sharpen the competitive edge for auction, through increasing the bidding density $d_i = \beta_i/\widetilde{a}_{i,t}$, which reflects the budget affordability of the AI user $u_i$ for acquiring one unit of edge inference resources to fulfill the user's edge inference resource request $\widetilde{a}_{i,t}$. Therefore, the AI user $u_i$ would minimize its edge inference resource request $\widetilde{a}_{i,t}$, associated with the DNN model partition request $\widetilde{s}_{i,t}$ to specify the DNN fragments offloaded for edge inference acceleration. In so doing, the personalized requirement of AI user $u_i$ on DNN inference latency and accuracy could be appropriately fulfilled in an on-demand manner. Thus, Phase I transforms the user-specific inference performance requirements into feasible bidding profiles (i.e., $\widetilde{s}_{i,t}$ and $\widetilde{a}_{i,t}$) that align with multi-dimensional decision-making variables $s_{i,t}$ and $a_{i,t}$ of problem **P1**.
- **Phase II (Consensus Estimate-based Competitive Auction):** Building on the result of edge inference demand analysis $\widetilde{s}_{i,t}$ and $\widetilde{a}_{i,t}$ for each AI user $u_i \in \mathcal{U}_t$ in Phase I, the AI service providers then jointly act as an auctioneer to finalize the multi-dimensional edge inference decision $\{p_t, \mathbf{S}_t, \mathbf{A}_t\}$ for problem **P1**. Note that, Phase II strategically adapts classic auction primitives by incorporating randomized consensus estimates and MoulinShenker cost-sharing techniques, thereby enabling the mechanism to satisfy the key desirable properties of Edge-AI market discussed in Section IV-C. The edge inference auction is meticulously designed in three steps, following the general principles outlined below:
  - ◇ *Revenue-Optimal Single-Price Omniscient Auction* (Step 1): The upper bound of revenue $\widetilde{\Pi}(\mathcal{U}_t)$ is obtained through simulating the revenue-optimal single-price omniscient auction [34]. This step provides an unattainable yet informative upper bound $\widetilde{\Pi}(\mathcal{U}_t)$ on the achievable revenue, serving as the optimal benchmark for the AERIA mechanism to approach at each billing cycle (time slot) $t$.
  - ◇ *Randomized Consensus Estimate Technique* (Step 2): A randomized consensus estimate function $f(\cdot)$ [35] is well-defined to protect against the misreporting behavior of any bidding AI user $u_i$, and to yield an approximate optimal target revenue $\widehat{\mathcal{U}}_t$. While ensuring *incentive compatibility*, this step theoretically guarantees that the target revenue $\widehat{\Pi}(\mathcal{U}_t)$ approximate the optimal revenue benchmark (i.e., the upper bound of revenue $\widetilde{\Pi}(\mathcal{U}_t)$ in Step 1) in *a good competitive ratio*.
  - ◇ *Cost Sharing Mechanism* (Step 3): The set of bid-winning AI users $\widehat{\mathcal{U}}_t$ is specified to contribute to the approximate optimal target revenue $\widehat{\Pi}(\mathcal{U}_t)$, based on the idea of Moulin-Shenker costing sharing mechanism [36]. The multi-dimensional decisions $\{p_t, \mathbf{S}_t, \mathbf{A}_t\}$ are finalized, where each bid-winning AI user $u_i \in \widehat{\mathcal{U}}_t$ is admitted to on-demand edge inference acceleration by following $s_{i,t} = \widetilde{s}_{i,t}$ and $a_{i,t} = \widetilde{a}_{i,t}$. The edge inference service is priced at a uniform rate $p_t$ (i.e. *non-discriminatory for different bidding AI users*). This uniform pricing facilitates the desirable property of *envy freeness* in the auction.

### *B. Phase I: Edge Inference Demand Analysis*

Phase I implements a fine-grained edge inference demand analysis to rigorously investigate the edge inference resource request $\widetilde{a}_{i,t}$ of each AI user $u_i$, in view of the following considerations:

- **Addressing the *self-interested nature of bidders*:** each bidding AI user $u_i$ is self-motivated to increase the bidding density $d_i = \beta_i/\widetilde{a}_{i,t}$, thereby promoting the competitiveness for auction. To this end, the AI user $u_i$ should minimize the amount of requested edge resources $\widetilde{a}_{i,t}$, on the premise of meeting its performance requirement on inference latency and accuracy.
- **Addressing the *edge resource usage efficiency*:** the AI service providers expect to take the most advantage of limited edge inference resources. Allocating edge inference resources in accordance with the minimum resource demand $\widetilde{a}_{i,t}$ allows for optimal utilization of the leased edge infrastructure, enabling accommodation of a greater number of AI users experiencing edge inference acceleration.
- **Addressing the *DNN inference workload balance*:** the computation capacity of end devices should be sufficiently utilized. Unless the AI user $u_i$ is powerless to satisfy the performance requirement of the DNN inference tasks, it would not request for the edge inference resource. Following this guideline, the DNN inference workload is appropriately distributed between edge and end devices in an on-demand manner.

A unique edge inference demand analysis problem **P2** is formulated for each AI user $u_i$, tailored to the user's specific local computation capacity $f_i$ and DNN inference latency requirement $t_i$. As an optimization subproblem to **P1**, the problem **P2** targets to figure out the optimal edge inference resource request $\widetilde{a}_{i,t}$ and the associated DNN model partition request $\widetilde{s}_{i,t}$. Here, $\widetilde{a}_{i,t}$ is defined in Eq. (10), indicating the minimum edge computation capacity (measured by FLOPS) requested by $u_i$ to fulfill its performance requirement on DNN inference. Note that $F_i^{edge}$ denotes the required computation overhead for accomplishing the edge inference task, while $t_i - T_i^{dev} - T_i^{net}$ represents the remaining available time for edge inference (excluding the local inference latency $T_i^{dev}$ and the network transmission latency $T_i^{net}$).

$$(\mathbf{P2}): \widetilde{a}_{i,t} = \min_{\widetilde{s}_{i,t}} \frac{F_i^{\text{edge}}(\widetilde{s}_{i,t})}{t_i - T_i^{\text{dev}}(\widetilde{s}_{i,t}) - T_i^{\text{net}}(\widetilde{s}_{i,t})} \tag{10}$$

$$\text{s.t.} \quad t_i - T_i^{\text{dev}}(\widetilde{s}_{i,t}) - T_i^{\text{net}}(\widetilde{s}_{i,t}) > 0 \tag{10a}$$

$$\widetilde{a}_{i,t} < f_e \tag{10b}$$

The terms $F_i^{\text{edge}}$, $T_i^{\text{dev}}$ and $T_i^{\text{net}}$ in the optimization objective function (10) are respectively defined in Eq. (3), Eq. (4) and Eq. (5). Each of these terms is governed by the exiting probability $\nu_j^k(\sigma)$ under the AI user $u_i$'s inference confidence criterion $\sigma_i$. Consequently, the optimization objective of problem **P2** inherently incorporates and respects the AI user $u_i$'s inference accuracy requirement. Constraint (10a) means that the device-edge synergistic inference latency $T_i$ should be less than $t_i$, suggesting the inference latency requirement of AI user $u_i$. Constraint (10b) implies that the edge inference resource request $\widetilde{a}_{i,t}$ of AI user $u_i$ does not exceed the computation capacity $f_e$ of edge cluster.

***Computation Complexity Analysis* (Phase I):** Through a trial-and-error process, each AI user $u_i$ iteratively evaluates $(g_{w_i}+1)$ candidate DNN model partition points (i.e., $\widetilde{s}_{i,t} = \{0, 1, ..., g_{w_i}\}$ where $g_{w_i}$ is the number of layers in the DNN model $M_{w_i}$) in problem **P2** to find the optimal edge inference resource request $\widetilde{a}_{i,t}$, therefore the computation complexity is $\mathcal{O}(g_{w_i})$. Furthermore, each AI user $u_i \in \mathcal{U}_t$'s problem **P2** is independent with no correlation, thus can be solved in parallel, which reduces the overall computational complexity of Phase I to $\mathcal{O}(\max_{u_i \in \mathcal{U}_t} g_{w_i})$.

### C. Phase II: Consensus Estimate-based Competitive Auction

Based on the edge inference demand analysis in Phase I, we further finalize the multi-dimensional decisions $\{p_t, \mathbf{S}_t, \mathbf{A}_t\}$ for problem **P1**, with the set of bid-winning AI users $\widehat{\mathcal{U}}_t$ at the time slot $t$ determined as well. Specifically speaking, a Consensus Estimate-based Competitive Auction (COCO) is proposed through combining three general thoughts, including 1) *Revenue-optimal single-price omniscient auction* [34], 2) *Randomized consensus estimate* [35], and 3) *Cost sharing mechanism* [36]. Correspondingly, the COCO auction is structured with three steps, which are Step 1, Step 2, and Step 3 as sketched in Section V-A and Fig. 2.

- **Step 1.** *Calculate the upper bound of revenue* $\widetilde{\Pi}(\mathcal{U}_t)$.

The Revenue-Optimal Single-price Omniscient Auction (ROSO) [34] is introduced to find out the upper bound of AI service providers' revenue $\widetilde{\Pi}(\mathcal{U}_t)$, as an optimal baseline to be approached at each billing cycle (time slot) $t$.

**Algorithm 1:** Revenue-Optimal Single-Price Omniscient Auction (**ROSO**)

**Input:**
DNN model partition point $\widetilde{s}_{i,t}$ of all $u_i \in \mathcal{U}_t$ by **P2**;
Edge inference resource demand $\widetilde{a}_{i,t}$ of all $u_i \in \mathcal{U}_t$ by **P2**.
**Output:**
Upper bound of revenue (omniscient-optimal) $\widetilde{\Pi}(\mathcal{U}_t)$;
Admitted AI users $\widetilde{\mathcal{U}}_t$;
Sum of requested edge inference resources $\widetilde{\Phi}_t$ from the admitted AI users $\widetilde{\mathcal{U}}_t$.

```
1  Initialize Ũ_t ← ∅, p_t^res ← (1 + γ) · p_t^r / f_e;
2  for each bidding AI user u_i ∈ U_t do
3      Calculate the bidding density of u_i, i.e., d_i ← β_i / ã_{i,t};
4      if d_i < p_t^res then
5          U_t ← U_t \ {u_i};
6  Order all u_i ∈ U_t by d_i in non-increasing rank;
7  Configure f'_e ← f_e;
8  for each bidding AI user u_i ∈ U_t do
9      if f'_e − ã_{i,t} > 0 then
10         f'_e ← f'_e − ã_{i,t};
11         Ũ_t ← Ũ_t ∪ {u_i};
12 Π̃(U_t) ← max_{u_i ∈ Ũ_t} (β_i / ã_{i,t}) · Σ_{k=1}^{i} ã_{k,t};
13 Φ̃_t ← argmax_{Σ_{k=1}^{i} ã_{k,t}} (β_i / ã_{i,t}) · Σ_{k=1}^{i} ã_{k,t};
14 return Π̃(U_t), Ũ_t, Φ̃_t;
```

***Definition* 1** (*Revenue-Optimal Single-price Omniscient Auction*, ROSO)**.** Let $\mathcal{U}_t$ ordered in non-increasing rank by the bidding density $d_i$, where $u_{[i]}$ represents the AI user who holds the $i$-th largest bidding density in $\mathcal{U}_t$. The ROSO auction aims to determine the user index $i$ such that $(\beta_i/\widetilde{a}_{i,t}) \cdot \sum_{k=1}^{i} \widetilde{a}_{k,t}$ is maximized, i.e.,

$$\widetilde{\Pi}(\mathcal{U}_t) = \max_{u_{[i]} \in \mathcal{U}_t} d_{[i]} \cdot \sum_{k=1}^{i} \widetilde{a}_{[k],t}, \tag{11}$$

where $\widetilde{\Pi}(\mathcal{U}_t)$ is the omniscient-optimal revenue, and the bidding density $d_i = \beta_i/\widetilde{a}_{i,t}$. All bidding AI users $u_j \in \mathcal{U}_t$ with $\beta_j \geq \beta_{[i]}/\widetilde{a}_{[i],t}$ win the ROSO auction, while the remaining AI users lose.

By Definition 1, Algorithm 1 elaborates the procedure of ROSO auction, where the omniscient-optimal revenue $\widetilde{\Pi}(\mathcal{U}_t)$ is calculated as the upper bound of revenue $\widetilde{\Pi}(\mathcal{U}_t)$. The AI service providers firstly preconfigure a reserve edge inference price[2] $p_t^{res} = (1+\gamma) \cdot p_t^r / f_e$ per the minimum profit-rate requirement of $\gamma$. Then, our ROSO auction calculates the bidding density $d_i$ of each AI user $u_i \in \mathcal{U}_t$, and eliminates the bidding users with $d_i < p_t^{res}$, such that the admitted AI users pay at least the reserve price $p_t^{res}$ (*Lines* 1-5).

[2]The reserve price $p_t^{res}$ suggests the lowest price that the AI service providers would accept from AI users to shield the minimum profit rate $\gamma$.

Afterwards,the set of admitted AI user $\widetilde{\mathcal{U}}_t$ by ROSO auction is determined, according to the non-increasing rank of the AI users' bidding densities $d_i$ (*Lines* 6-11). The AI user $u_i$ will be denied edge inference services, in the case of insufficient edge inference resources to meet its demand $\widetilde{a}_{i,t}$; otherwise, the AI user $u_i$ will be admitted with the amount of available edge resources correspondingly updated (*Lines* 9-11). Finally, our ROSO auction calculates the upper bound of revenue (omniscient-optimal) $\widetilde{\Pi}(\mathcal{U}_t)$ based on Eq. (11), which will be transferred to Step 2 of our AERIA mechanism (*Lines* 12-14).

Note that, the ROSO auction has perfect knowledge of each bidding AI user's market profile. Hence, the ROSO auction is not incentive compatible, that is, the bidding AI user can misreport the edge inference bid to earn the auction. Owing to incomplete market profile information, we must introduce appropriate relaxation for revenue optimality. As a result, the question arises as to how an incentive compatible auction is developed while approaching the upper bound of revenue $\widetilde{\Pi}(\mathcal{U}_t)$.

- **Step 2.** *Randomly assign the target revenue* $\widehat{\Pi}(\mathcal{U}_t)$ *within a credible interval.*

To ensure the incentive compatibility for edge inference auction, we pursue such *a bid-independent auction* where the edge inference price is independent of the bidding profiles reported by the bidding AI users. In other words, the auction outcomes $\{p_t, \mathbf{S}_t, \mathbf{A}_t\}$ remain unaffected, regardless of whether the AI user $u_i$ truthfully reports the bidding budget $\beta_i$. To this end, the notion of *Consensus Estimate* [35] is introduced to stabilize the revenue which cannot be altered by faithless AI users.

***Definition* 2** (*Consensus Estimate*)**.** Let $\mathcal{U}_t^{-i} = \mathcal{U}/\{u_i\}$ denote the set of all AI users $\mathcal{U}_t$ excluding $u_i$. For a given $\delta$, a function $f(\cdot)$ is a *$\delta$-consensus estimate of* $\widetilde{\Pi}(\mathcal{U}_t)$, if: $\Pi(\mathcal{U}_t^{-i}) \in [\Pi(\mathcal{U}_t)/\delta, \Pi(\mathcal{U}_t)] \Rightarrow f(\Pi(\mathcal{U}_t)) = f(\Pi(\mathcal{U}_t^{-i})) \approx \Pi(\mathcal{U}_t)$.

By Definition 2, the consensus estimate $f(\Pi(\mathcal{U}_t))$ yields a sufficiently accurate estimate of upper-bounded revenue $\widetilde{\Pi}(\mathcal{U}_t)$, i.e., $f(\Pi(\mathcal{U}_t)) \approx \Pi(\mathcal{U}_t)$. More importantly, the consensus estimate $f(\Pi(\mathcal{U}_t))$ for the upper-bounded revenue $\widetilde{\Pi}(\mathcal{U}_t)$ holds immunity to the misreporting behavior of any bidding AI user $u_i \in \mathcal{U}_t$, which facilitates the *incentive compatibility*. That is, the consensus estimate $f(\Pi(\mathcal{U}_t))$ is aligned with the estimate of $\Pi(\mathcal{U}_t^{-i})$ for each AI user $u_i \in \mathcal{U}_t$, i.e., $f(\Pi(\mathcal{U}_t)) = f(\Pi(\mathcal{U}_t^{-i}))$.

Also note that, a consensus estimate for $\Pi(\mathcal{U}_t)$ could be secured and formulated by $f(\Pi(\mathcal{U}_t))$ if $\Pi(\mathcal{U}_t^{-i})$ is limited within $[\Pi(\mathcal{U}_t)/\delta, \Pi(\mathcal{U}_t)]$. Therefore, the parameter $\delta$ for achieving such a consensus estimate needs to be well specified.

**PROPOSITION** 1. Suppose that any one of edge inference bids by AI users $\mathcal{U}_t$ is eliminated, then the upper bound of revenue $\widetilde{\Pi}(\mathcal{U}_t)$ can be reduced at most a factor of $(\widetilde{\Phi}_t - \zeta)/\widetilde{\Phi}_t$, where $\zeta = \max(\widetilde{a}_{1,t}, \widetilde{a}_{2,t}, ..., \widetilde{a}_{N(t),t})$.

*Proof.* See Appendix A of the supplementary material. ☐

According to Proposition 1, we can define $\delta = \widetilde{\Phi}_t/(\widetilde{\Phi}_t - \zeta)$. Thus, $\Pi(\mathcal{U}_t)/\delta \le \Pi(\mathcal{U}_t^{-i}) \le \Pi(\mathcal{U}_t)$ holds for all $u_i \in \mathcal{U}_t$, promising that $\Pi(\mathcal{U}_t^{-i})$, $\forall u_i \in \mathcal{U}_t$ is always within a constant fraction of $\Pi(\mathcal{U}_t)$.

---

**Algorithm 2:** Consensus-Estimate-based Target Revenue Settling Algorithm (**CENTRE**)

---

**Input:**
Edge inference resource demand $\widetilde{a}_{i,t}$ of all $u_i \in \mathcal{U}_t$ by **P2**;
Upper bound of revenue (omniscient-optimal) $\widetilde{\Pi}(\mathcal{U}_t)$;
Admitted AI users $\widetilde{\mathcal{U}}_t$ by ROSO auction;
Sum of requested edge inference resources $\widetilde{\Phi}_t$ from the admitted AI users $\widetilde{\mathcal{U}}_t$ by ROSO auction.

**Output:**
Target revenue $\widehat{\Pi}(\mathcal{U}_t)$.

Configure $\delta \leftarrow \widetilde{\Phi}_t/(\widetilde{\Phi}_t - \zeta)$, where $\zeta = \max_{u_i \in \widetilde{\mathcal{U}}_t} \widetilde{a}_{i,t}$;
Find $y \leftarrow \arg\max_{y} \frac{1}{\ln y}\left(\frac{1}{\delta} - \frac{1}{y}\right)$;
**repeat**
  Generate the random param. $\varepsilon$ uniformly from $[0, 1]$;
  Update the target revenue as $\widehat{\Pi}(\mathcal{U}_t) \leftarrow y^{\lfloor \log_y \widetilde{\Pi}(\mathcal{U}_t) - \varepsilon \rfloor + \varepsilon}$;
**until** $\widehat{\Pi}(\mathcal{U}_t) \le \widetilde{\Pi}(\mathcal{U}_t)/\delta$;
**return** $\widehat{\Pi}(\mathcal{U}_t)$;

---

Note that, no deterministic function $f(\cdot)$ is a *$\delta$-consensus estimate of* $\Pi > 0$. A simple induction by contradiction reveals that, $f(\cdot)$ would have to be a constant function, assuming the existence of such a deterministic function $f(\cdot)$ to be consensus for $\Pi > 0$. However, the constant which is lower-bounded on $\Pi > 0$ has to be non-positive, which intuitively contradicts with the requirement $f(\cdot) > 0$ for revenue estimation. Following the above analysis, a randomized consensus estimate function $f(\cdot)$ is constructively defined as follows.

***Definition* 3** (*Randomized Consensus Estimate Function*)**.** Given the parameter $y > \delta$ chosen to maximize the quality of revenue estimate [3], and $\varepsilon$ picked uniformly on $[0, 1]$, the randomized consensus estimate function $f(\varpi)$ for any $\varpi \in [\Pi(\mathcal{U}_t)/y, \Pi(\mathcal{U}_t)]$ is formally defined as

$$f(\varpi) = y^{\lfloor \log_y \varpi - \varepsilon \rfloor + \varepsilon}, \tag{12}$$

which formulizes $\varpi$ rounded down to the nearest $y^{w+\varepsilon}$ for $w \in \mathbb{Z}$.

To examine whether a consensus estimate is reached by the randomized function $f(\varpi)$, we then derive a judging criteria $f(\Pi(\mathcal{U}_t)) \le \Pi(\mathcal{U}_t)/\delta$ in Theorem 1. Before that, we preliminarily draw an important conclusion in Lemma 1.

**LEMMA** 1. *The randomized function* $f(\varpi)$ *is a consensus estimate for* $u_i \Leftrightarrow f(\Pi(\mathcal{U}_t)) \le \Pi(\mathcal{U}_t^{-i})$.

*Proof.* See Appendix B of the supplementary material. ☐

**THEOREM** 1. *The randomized function* $f(\varpi)$ *is a consensus estimate for all AI users* $u_i \in \mathcal{U}_t$*, if* $f(\Pi(\mathcal{U}_t)) \le \Pi(\mathcal{U}_t)/\delta$.

*Proof.* See Appendix C of the supplementary material. ☐

Next, we will investigate the competitiveness against the upper-bounded revenue $\Pi(\mathcal{U}_t)$ achieved by the randomized consensus estimate function $f(\varpi)$, prior to which the probabilistic distribution of $f(\varpi)$ is derived in Lemma 2.

[3] Quality of revenue estimate is assessed by how good a competitive ratio is reached in revenue maximization, as detailed later in Theorem 2.

**LEMMA** 2. *The randomized function $f(\varpi)$ is distributed identically to $\varpi \cdot y^{\varepsilon'-1}$ for $\varepsilon'$ chosen uniformly on $[0,1]$.*

*Proof.* See Appendix D of the supplementary material. □

**THEOREM** 2 (*Competitiveness in Revenue Maximization*). $\mathbf{E}\left[f\left(\Pi\left(\mathcal{U}_t\right)\right)\right] \geq \frac{\widetilde{\Pi}(\mathcal{U}_t)}{\ln y}\left(\frac{1}{\delta} - \frac{1}{c}\right)$.

*Proof.* See Appendix E of the supplementary material. □

From Theorem 2, we can consider the expected value of $f\left(\Pi\left(\mathcal{U}_t\right)\right)$ which approaches the upper-bounded revenue $\widetilde{\Pi}(\mathcal{U}_t)$ with a good competitive ratio $\frac{1}{\ln y} \cdot \left(\frac{1}{\delta} - \frac{1}{y}\right)$. For the choice of parameter $y$, we configure the value of $y$ that maximizes $\mathbf{E}\left[f\left(\Pi\left(\mathcal{U}_t\right)\right)\right]/\Pi\left(\mathcal{U}_t\right)$, given a fixed $\delta$. In so doing, a better bound for $\mathbf{E}\left[f\left(\Pi\left(\mathcal{U}_t\right)\right)\right]$ is guaranteed, suggesting a high-quality revenue estimate.

Algorithm 2 describes the <u>C</u>onse<u>n</u>sus-Estimate-based <u>T</u>arget <u>R</u>evenue S<u>e</u>ttling procedure (CENTRE). Owing to the delicate design of randomized consensus function $f\left(\varpi\right)$ (see *Theorem* 1 and *Theorem* 2), we can adopt the consensus estimate value $f\left(\Pi\left(\mathcal{U}_t\right)\right)$ as the target revenue $\widehat{\Pi}(\mathcal{U}_t)$. Thereinto, the target revenue $\widehat{\Pi}(\mathcal{U}_t)$ is randomly assigned within a credible interval $[\log_y \Pi(\mathcal{U}_t) - 1, \widetilde{\Pi}(\mathcal{U}_t)/\delta]$. In other words, the essence of CENTRE is to suffice the randomized consensus function $f\left(\varpi\right)$ until meeting the judging criteria $f\left(\Pi\left(\mathcal{U}_t\right)\right) \leq \Pi\left(\mathcal{U}_t\right)/\delta$ (*Lines* 3-6), such that our AERIA mechanism can be both *incentive compatible* and *competitive in revenue maximization*.

- **Step 3.** *Collect the target revenue $\widehat{\Pi}(\mathcal{U}_t)$ from bid-winning AI users $\widehat{\mathcal{U}}_t$.*

Following the previous Step 2, we finalize the multi-dimensional decisions $\{p_t, \mathbf{S}_t, \mathbf{A}_t\}$ to collect the target revenue $\widehat{\Pi}(\mathcal{U}_t)$, which is based on the Moulin-Shenker costing sharing mechanism [36]. The set of bid-winning AI users is specified as $\widehat{\mathcal{U}}_t$ for target-revenue extraction.

The <u>R</u>evenue <u>E</u>xtraction <u>Al</u>gorithm (REAL) based on the cost-sharing mechanism is elaborated in Algorithm 3. The target revenue $\widehat{\Pi}(\mathcal{U}_t)$ is collected from the bid-winning AI users $\widehat{\mathcal{U}}_t$. In details, REAL firstly derives the set of bid-winning AI users $\widehat{\mathcal{U}}_t$ through iteratively excluding from $\widetilde{\mathcal{U}}_t$ the underbudgeted AI users whose bidding density $d_i = \beta_i/\widetilde{a}_{i,t}$ falls below the provisional edge inference price $p'_t$ (*Lines* 4-10). Note that the provisional edge inference price $p'_t$ is synchronously updated following the updated $\widehat{\Phi}_t$ in each iteration. Having determined the bid-winning AI users $\widehat{\mathcal{U}}_t$ and the corresponding $\widehat{\Phi}_t = \sum_{u_i \in \widehat{\mathcal{U}}_t} \widetilde{a}_{i,t}$ after iterations, REAL finalizes the edge inference price as $p_t \leftarrow p'_t$. The edge inference price is set according to $p_t = \widehat{\Pi}(\mathcal{U}_t)/\widehat{\Phi}_t$, where $\widehat{\Phi}_t$ represents the sum of requested edge inference resources from the bid-winning AI users $\widehat{\mathcal{U}}_t$.

With multi-dimensional decisions on $\{p_t, \mathbf{S}_t, \mathbf{A}_t\}$ finalized, each bidding-winning user $u_i \in \widehat{\mathcal{U}}_t$ possesses a bidding density $d_i \geq p_t$ to afford the payment of $p_t \cdot \widetilde{a}_{i,t}$. This ensures a *budget-balanced* auction, where the underbudgeted AI users whose bidding density $d_i$ fails $p_t$ have been eliminated in prior iterations (*Lines* 4-10). Meanwhile, each bid-winning AI user $u_i$ is pleased with its own experienced edge inference service in terms of inference accuracy and latency, while the edge inference service is valued at a flat price $p_t$ (i.e. *non-discriminatory for different bidding AI users*). In so doing, the desirable property of *envy freeness* is facilitated in auction.

---

**Algorithm 3:** <u>R</u>evenue <u>E</u>xtraction <u>Al</u>gorithm based on the Cost-Sharing Mechanism (**REAL**)

**Input:**
DNN model partition point $\widetilde{s}_{i,t}$ of all $u_i \in \mathcal{U}_t$ by **P2**;
Edge inference resource demand $\widetilde{a}_{i,t}$ of all $u_i \in \mathcal{U}_t$ by **P2**;
Target revenue $\widehat{\Pi}(\mathcal{U}_t)$ by CENTRE algorithm;
Admitted AI users $\widetilde{\mathcal{U}}_t$ by ROSO auction;
Sum of requested edge inference resources $\widetilde{\Phi}_t$ from the admitted AI users $\widetilde{\mathcal{U}}_t$ by ROSO auction.

**Output:**
Set of bid-winning AI users $\widehat{\mathcal{U}}_t$;
Edge inference price per unit FLOPS $p_t$;
Edge inference resource allocation vector $\mathbf{A}_t = (a_{1,t}, ..., a_{N_t,t})$;
DNN model partition decision vector $\mathbf{S}_t = (s_{1,t}, ..., s_{N_t,t})$.

1 Initialize $\mathbf{A}_t \leftarrow (0, .., 0)$, and $\mathbf{S}_t \leftarrow (\text{NULL}, .., \text{NULL})$;
2 Configure $\widehat{\mathcal{U}}_t \leftarrow \widetilde{\mathcal{U}}_t$, $\widehat{\Phi}_t \leftarrow \widetilde{\Phi}_t$, and $p'_t \leftarrow \widehat{\Pi}(\mathcal{U}_t)/\widetilde{\Phi}_t$;
3 Order all $u_i \in \widetilde{\mathcal{U}}_t$ by the bidding density $d_i$ in a non-increasing rank;
4 **for** *each bidding AI user* $u_i \in \widetilde{\mathcal{U}}_t$ **do**
5     **if** $d_i < p'_t$ **then**
6         $\widehat{\Phi}_t \leftarrow \widehat{\Phi}_t - \widetilde{a}_{i,t}$;
7         $\widehat{\mathcal{U}}_t \leftarrow \widehat{\mathcal{U}}_t \backslash \{u_i\}$;
8         $p'_t \leftarrow \widehat{\Pi}(\mathcal{U}_t)/\widehat{\Phi}_t$;
9     **else**
10         **break**;
11 $p_t \leftarrow p'_t$; ▷ Edge Inference Price
12 **for** *each bid-winning AI user* $u_i \in \widehat{\mathcal{U}}_t$ **do**
13     $a_{i,t} \leftarrow \widetilde{a}_{i,t}$; ▷ Edge Inference Resource Allocation
14     $s_{i,t} \leftarrow \widetilde{s}_{i,t}$; ▷ DNN model partition
15 **return** $\widehat{\mathcal{U}}_t$, $p_t$, $\mathbf{A}_t$, and $\mathbf{S}_t$;

---

***Summary*** **(Phase II):** Owing to the exquisite design on randomized consensus estimate, the COCO auction realizes the optimization objective of problem **P1** by finding out an approximate optimal target revenue $\widehat{\mathcal{U}}_t$ with a good competitive ratio. To present the *competitiveness in revenue maximization*, the target revenue $\widehat{\mathcal{U}}_t$ is compared with the upper-bounded revenue $\widetilde{\Pi}(\mathcal{U}_t)$ through competitive analysis. Meanwhile, the target revenue $\widehat{\mathcal{U}}_t$ endorsed by consensus estimate is bid-independent (i.e., *immune to the misreported bid*), thereby facilitating the *incentive compatibility*. Finally, the cost sharing mechanism enables the target revenue $\widehat{\mathcal{U}}_t$ collected at a flat price $p_t$ (i.e. *non-discriminatory for different bidding AI users*), which ultimately leads to an *envy-free* auction. In this way, a fair and truthful trading environment for edge inference services is fostered in the Edge-AI market.

***Computation Complexity Analysis*** **(Phase II):** The overall computation complexity of COCO auction derives from the following three steps.

- **Step 1** (*Algorithm* 1 - **`ROSO`**)**:** The sort-up operation of $N(t)$ bidding AI users contributes to the majority of computation overhead, resulting in the computation complexity of $\mathcal{O}(N(t) \cdot \log N(t))$.
- **Step 2** (*Algorithm* 2 - **`CENTRE`**)**:** Suppose that $C_2$ iterations (*lines* 3-6) are required until reaching the consensus estimate criteria $\widehat{\Pi}(\mathcal{U}_t) \leq \widetilde{\Pi}(\mathcal{U}_t)/\delta$, then the computation complexity of this step is $\mathcal{O}(C_2)$.
- **Step 3** (*Algorithm* 3 - **`REAL`**)**:** The admitted AI users $\widetilde{U}_t$ by **`ROSO`** auction is ordered according to the bidding density $d_i$ (*line* 3), which accounts for the majority of computation overhead in Step 3. Thus, the computation complexity of this step is $\mathcal{O}(\widetilde{N}(t) \cdot \log \widetilde{N}(t))$, where $\widetilde{N}(t) =\mid \widetilde{U}_t \mid$.

To summarize, the overall computation complexity is $\mathcal{O}(N(t) \cdot \log N(t) + C_2 + \widetilde{N}(t) \cdot \log \widetilde{N}(t))$ for Phase II.

## VI. Performance Evaluation

### A. Experimental Setup

**Multi-exit DNN Model and Inference Dataset Benchmarks:** We select four representative pre-trained ME-DNN models for device-edge synergistic inference, emulating the diverse AI services offered by multiple AI service providers. All ME-DNN models follow a common 80:20 split between training and test datasets. In addition to the original final exit, each pre-trained ME-DNN model contains three early-exit branches that equally apportion the computation overhead of main branch, termed shallow exit, middle exit, and deep exit. The DNN inference tasks are configured according to the following four scales:

- ***Giant-Scale Tasks*****:** With 30,016 German-English sentence pairs for machine translation, the multi-exit BART (ME-BART) [37] is pre-trained and inferred on the large-scale multilingual multimodal Multi30K dataset [38]. It is a representative large language model (LLM) with an encoder-decoder transformer architecture, widely used for generative AI tasks.
- ***Large-Scale Tasks*****:** With 1,000 object classes and 50,176 colorful pixels per image, the multi-exit ResNet-34 (ME-ResNet-34) [26] is pre-trained and inferred on the ImageNet-1K dataset [39].
- ***Medium-Scale Tasks*****:** With 100 object classes and 1,024 colorful pixels per image, the multi-exit VGG-16 (ME-VGG-16) [40] is pre-trained and inferred on the CIFAR-100 dataset [41].
- ***Small-Scale Tasks*****:** With 10 object classes and 1,024 colorful pixels per image, the multi-exit AlexNet (ME-AlexNet) [39] is pre-trained and inferred on the CIFAR-10 dataset [41].

**AI Users (End Devices):** The computation capacity $f_i$ of each end device $u_i$ is drawn from the uniform distribution across [0.5, 5] GFLOPS. At each time slot $t$, the edge inference bid placed by each AI user $u_i$ is $R_i = \{\beta_i, t_i, \sigma_i, w_i\}$, where $w_i$ randomly derives from our selective ME-DNN models, and $\beta_i, t_i, \sigma_i$ are user-specific. The geographical distribution of all AI users (end devices) originates from the Shanghai Telecom dataset[4]. It records the trajectory of 9,481 mobile users in Shanghai, China across 15 days, in accordance with our concerned device-edge scenario. To simplify the trace-driven experiment, we randomly select a fraction of 1,500 mobile users from the entire dataset. The selected mobile users dynamically determine to place edge inference bids across time slots. As shown on the left side of Fig. 5, there are approximately 39 to 112 edge inference bids placed at each time slot. The duration $\tau$ of a time slot is set as one hour, per the minimum billing cycle of Microsoft Azure [42].

**Edge Cluster** $\boldsymbol{e}$**:** It is geographically deployed at the centroid of AI user (end device) activity area. The computation capacity $f_e$ of edge server is configured as 3,930 GFLOPS, which matches with the performance of three sets of Intel Core i7-13700K machines. The energy cost to operate the edge cluster $e$ is estimated as 1.2 × 375 = 450 Watts, in which the ratio of power usage effectiveness (PUE) is set as 1.2 based on [43]. To assess the volatile rental/operational cost of edge infrastructures, as shown on the right side of Fig. 5, we adopt the trajectory of Ontario's hourly power prices[5] for 15 days (i.e., from September 6 to September 20, 2022) in the trace-driven experiment. Otherwise, the electricity price $p_t^e$ is set as 0.1056 $\$/kWh$, that is the median of our adopted Ontario's power price trajectory.

**Wireless Network:** The average propagation delay $\xi_i$ is related to the communication distance $\triangle_{i,e}$ between the edge cluster $e$ and the end device $u_i$, i.e., $\xi_i = \triangle_{i,e}/c$ (speed of light). Unless otherwise specified, the wireless data rate $r_{i,t}$ between edge cluster $e$ and end device $u_i$ is drawn from the uniform distribution over the stable range of [20, 30] Mbps. To further evaluate the impact of real-world wireless network volatility, we conduct a trace-driven experiment based on 4G/LTE bandwidth measurements collected in the urban area of Ghent, Belgium[6]. As shown by the middle of Fig. 5, the wireless data rate $r_{i,t}$ fluctuates across 360 time slots within a wide range of [7, 60] Mbps. The lower-data-rate conditions along this trajectory capture the effects of communication volatility including packet loss and signal fading.

### B. Performance Benchmarks

Our proposed **`AERIA`** mechanism is compared against four representative benchmark approaches, including three state-of-the-art algorithms (i.e., *IAO* [8], *Edgent* [11], and *AMR*[2] [44]), and a *fixed-profit-rate* baseline approach.

- **`IAO`** [8]**:** It performs model partition on the vanilla DNN and heuristically determines the edge inference resource demand for AI users. The AI user $u_i$ with the highest inference latency is prioritized through being allocated with the most edge inference resources, thereby optimizing the overall inference performance across all users. The edge inference price is strategically determined to optimize the AI service providers' revenue, following a pricing approach similar to AERIA.
- **`Edgent`** [11]**:** It conducts DNN model partition and right-sizing by configuring an exit branch, where the DNN inference process stops early. The exit branch is preferably chosen at a deeper layer, subject to each AI user's inference latency requirement. Multiple AI users compete for the finite inference resources on edge servers, and a subset of bid-winning users, who enhance the AI service providers' revenue, ultimately secure edge inference resources.

[4]https://wangshangguang.github.io/telecom_dataset/.

[5]http://www.ieso.ca/power-data.

[6]https://users.ugent.be/~jvdrhoof/dataset-4g/

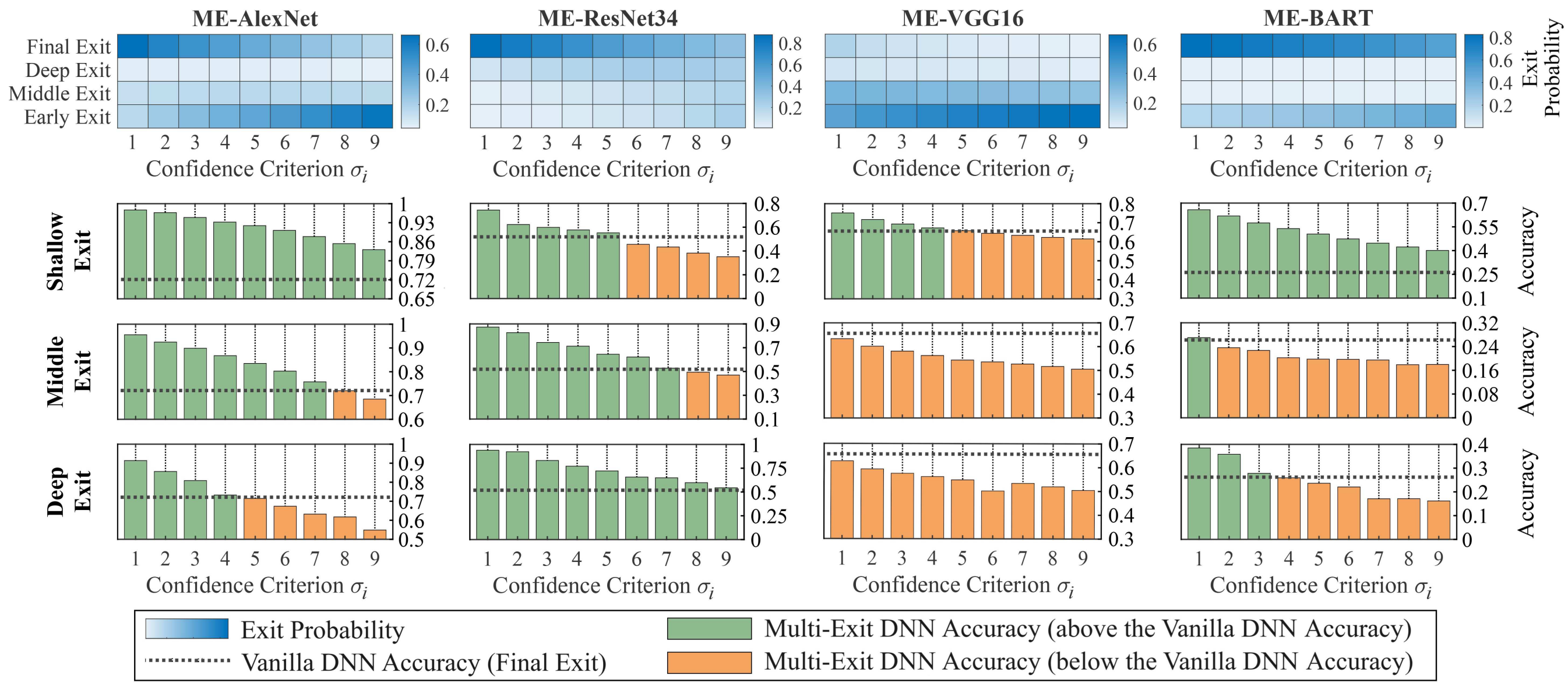

Fig. 3: Impact of Confidence Criterion $\sigma_i$ on Inference Accuracy and Exiting Probability at Different Early-Exit Branches.

- **`AMR`**[2] [44]**:** It performs layer-wise pruning on the vanilla DNN for each AI user to ensure that all their edge inference requests can be efficiently processed within the computation capacity of edge servers. The pruned DNN is preferably as deep as possible, assuming that a deeper DNN provides better inference accuracy. To ensure affordability for all bidding AI users, the edge inference price is set to the lowest bidding density among them.
- **`Fixed Profit Rate = 1.0`:** Differing from `AERIA` for revenue maximization, it takes a rigid earning target for the AI service providers with the pre-defined profit-rate objective of 1.0.

All the trace-driven experimental results about the `AERIA`, `IAO`, `Edgent`, and `Fixed Profit Rate = 1.0` are averaged over 120 runs.

### *C. Numerical Experiments*

**Impact of Confidence Criterion on ME-DNN Accuracy:** Fig. 3 demonstrates the inference accuracy and exit probability at different early-exit branches for ME-AlexNet, ME-ResNet34, ME-VGG16, and ME-BART models, under various confidence-criterion settings. Formally, we define inference accuracy as Inf. Acc. $= \sum_{s \in \text{Test}} \text{AccScore}_s$, where $\text{AccScore}_s$ denotes the accuracy score for each sample $s$ in the test dataset. Technically speaking, a more strict confidence criterion $\sigma_i$ (i.e., lower-valued) implies more stringent requirement on inference accuracy. That is, the inference accuracy generally increases along with a more strict confidence criterion. Meanwhile, if the entropy of exit output falls below the pre-configured confidence criteria $\sigma_i$, the DNN inference task can thereby exit early. As a result, more DNN inference tasks will be terminated at the shallow exit if relaxing its confidence criteria $\sigma_i$. Accordingly, the exit probability at middle and deep exits will remain relatively low, especially when the confidence criteria $\sigma_i$ is loose.

**Multi-exit DNN *v.s.* Vanilla DNN:** Compared with the corresponding vanilla DNNs, our ME-AlexNet, ME-ResNet34, and ME-VGG16 models respectively have the average accuracy loss of 1.97%, 2.47%, and 3.77%. Even better, the ME-BART model generally offers a slight accuracy[7] improvement of 1.10% compared to the vanilla BART model. It can be noticed that the inference accuracy of ME-DNN models surprisingly outperforms that of vanilla DNNs in some cases. Exemplified by the ME-AlexNet, its accuracy increase reaches up to 1%-8% at certain early-exit branches, subject to confidence criterion. Such a observation suggests that the ME-DNN inference is capable of mitigating the general vulnerability of DNNs termed *overethinking* [47]. That is, DNNs tend to suffer from redundant computation overhead on some simple-to-infer samples, hence resulting in overfitting inference and accuracy reduction. Heterogenous ovethinking situations exists for different DNNs, which depends on differentiated model structures. As demonstrated in Fig. 3, accuracy improvement can be reached at the shallow exit for ME-AlexNet, ME-VGG16 and ME-BART models, while ME-ResNet34 prefers the deep exit. To summarize, the adoption of ME-DNNs demonstrates insignificant affect on inference accuracy, even alleviating the overthinking issue of DNNs in some cases.

**Efficacy on Generative AI Tasks:** We pre-train an ME-BART model, a LLM for generative AI tasks, on the Multi30K dataset for machine translation. The right side of Fig. 3 illustrates the inference accuracy and exit probability of ME-BART model. With a looser confidence-criterion setting of $\sigma_i$, a greater portion of text samples, ranging from 18% to 48%, are qualified to complete inference early. Especially, the phenomenon of "ovethinking" [47] is more apparent for the ME-BART model, due to its extremely large model that is inherently overfitting prone for simple-to-infer samples. Owing to the multi-exit inference design, an average accuracy improvement of 1.10% is achieved for the ME-BART model, in comparison with the vanilla BART model. These results highlight that our edge inference design effectively addresses advanced LLM inference tasks, and is applicable to broader generative AI scenarios.

[7] We adopt the recognized BiLingual Evaluation Understudy (BLEU) standard [45][46] to evaluate the inference accuracy of ME-BART model.

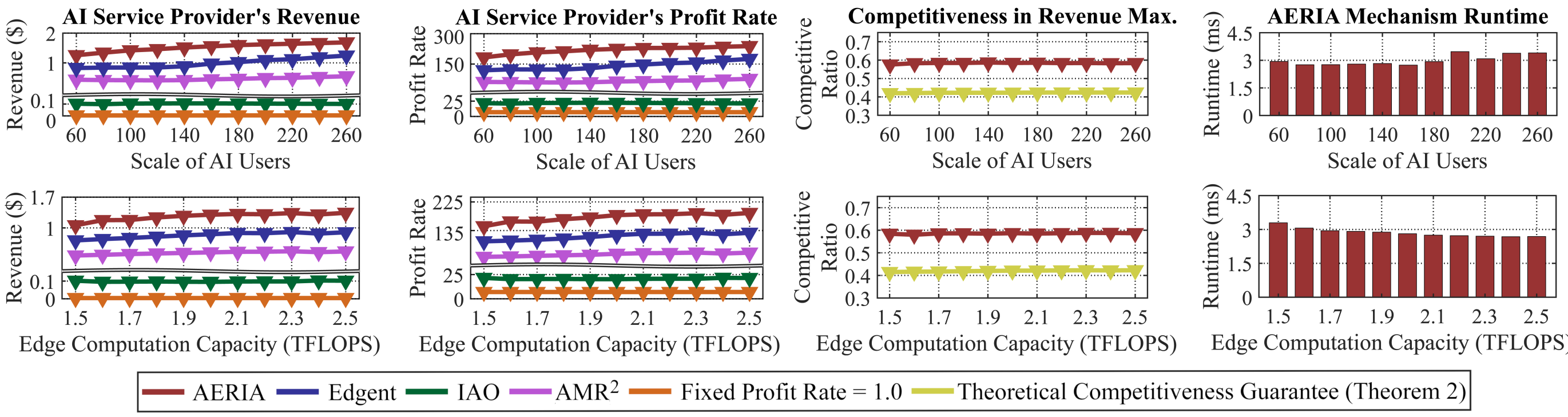

Fig. 4: Impact of System Scale on Revenue Optimality and Algorithmic Efficiency.

**Impact of System Scale on Revenue Optimality:** In Fig. 4, the revenue optimality of our `AERIA` mechanism is demonstrated under different scales of AI users $N_t$, and edge computation capacities $f_e$. Compared with performance benchmarks, our proposed `AERIA` mechanism realizes the highest revenue/profit rate for the AI service providers, which coincides with their optimization objective of revenue maximization. Meanwhile, the revenue/profit rate earned by our `AERIA` mechanism increases with the scale of AI users $N_t$ and edge computation capacity $f_e$. Intuitively, the AI service providers can gain higher revenue from the greater scale of AI users. It suggests that more AI users can be accommodated at an edge clusterv $e$ with greater computation capacity, thereby acquiring higher revenue/profit rate. Nonetheless, the revenue and profit rate of `IAO` remain almost unchanged at low level, regardless of different scales of AI users $N_t$, and edge computation capacities $f_e$. This is because the `IAO` algorithm, which neglects the budget affordability of each bidding AI user, has to decline a major portion of edge inference requests whose bidding budget is insufficient. The `Fixed Profit Rate = 1.0` approach strictly follows its target profit rate, hence gaining the lowest revenue/profit rate.

The actual competitiveness ratio achieved by our `AERIA` mechanism is also compared with the baseline of theoretical competitiveness guarantee for revenue maximization (as detailed in Theorem 2). By definition, the competitiveness ratio refers to the ratio between the actual revenue (i.e., target revenue) $\widehat{\Pi}(\mathcal{U}_t)$ and upper-bounded revenue $\widetilde{\Pi}(\mathcal{U}_t)$. Regardless of different scales of AI users $N_t$, and edge computation capacities $f_e$, the actual competitiveness ratio always holds steady and much higher than the theoretical baseline, which implies the validity of Theorem 2.

**Impact of System Scale on Algorithmic Efficiency:** We evaluate the execution efficiency of our proposed `AERIA` mechanism by measuring its runtime under various system scales. The runtime measurement is performed on a machine equipped with an Intel Core i9-10920X @3.50GHz. As shown on the right side of Fig. 4, the runtime of `AERIA` is always around 3 milliseconds, suggesting its high scalability across different system scales. The efficient execution demonstrates the feasibility of our `AERIA` for synergistic device-edge inference.

### *D. Real-World Trace-Driven Experiments*

Considering the system volatility in terms of edge resources and requests, we conduct real-world trace-driven experiments with varying edge inference bids, wireless data rates and hourly electricity prices. In general, more edge inference bids placed at each time slot imply greater edge inference demand, thereby incurring higher revenue and profit rate for the AI service providers. Higher wireless data rates would facilitate the synergistic inference between the edge cluster and end devices, thereby incentivizing more AI users to pay for edge inference services, leading to higher revenue/profit rate earned by the AI service providers. A higher hourly electricity price means that the rental/operational cost of edge infrastructures increases, hence reducing the earned revenue/profit rate for the AI service providers.

Under dynamic market environments, our `AERIA` mechanism always gains higher revenue compared to the four representative baselines, demonstrating that AERIA can effectively handle the dynamic nature of the Edge-AI market. Differing from probabilistic multi-exit setting in `AERIA`, `Edgent` straightforwardly right-sizes the vanilla DNN via early exit, thereby demanding for less edge inference resources to undertake the right-sized DNN calculation. Hence, `Edgent` has to incentivize the edge inference resource demand by configuring a lower edge inference price, resulting in less revenue earnings. `AMR`$^2$ enrolls all bidding AI users to secure edge inference resources and be served at the edge. To achieve this, the edge inference price must be set affordably for all AI users (i.e., as the lowest bidding density among them), which consequently results in much lower profit rate and revenue earnings. Since `IAO` ignores the budget affordability while determining the edge inference resource demand, the AI service providers have to reject the edge inference requests whose bidding budget is insufficient. As a result, the earned revenue by `IAO` is much lower than `AERIA` and `Edgent`. In terms of `Fixed Profit Rate = 1.0` approach, it gains the lowest revenue due to its rigid algorithmic design. The `Fixed Profit Rate = 1.0` approach always keeps its target profit rate across time slots.

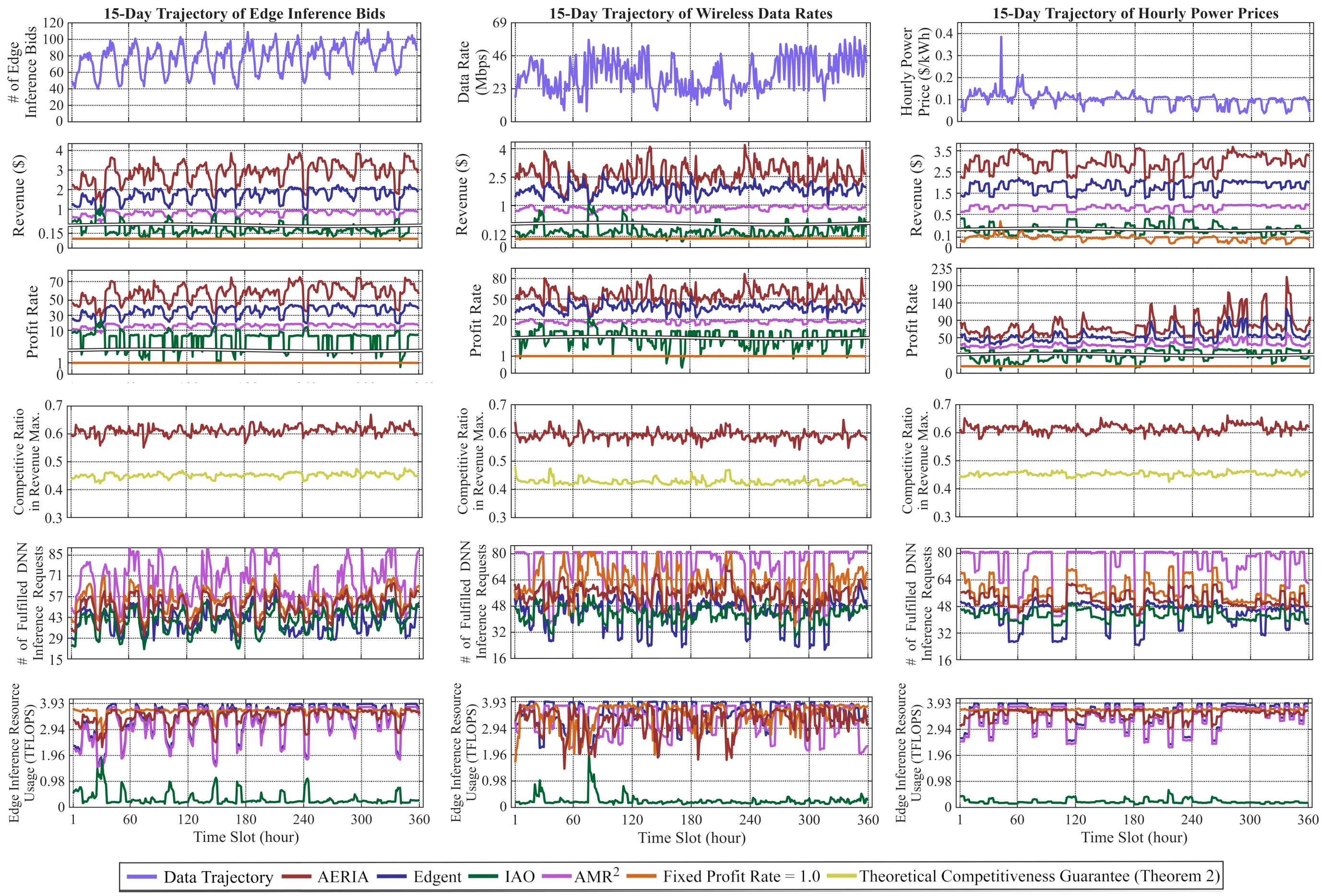

Fig. 5: Comparative Simulation Results based on Various Real-World Trace Data.

Similar to Fig. 4, Fig. 5 evaluates the actual competitiveness ratio achieved by our `AERIA` mechanism for revenue maximization in real-world trace-driven experiments. It can be observed that the actual competitiveness ratio is always much higher than the baseline of theoretical competitiveness guarantee for revenue maximization, which suggests the efficacy of Theorem 2. By addressing the personalized inference requirements of different AI users, our `AERIA` mechanism also consistently ranks among the top in both user accuracy and latency adaptation, as reflected in the number of fulfilled DNN inference requests. Meanwhile, the edge inference resource utilization of `AERIA` achieves a leading level as in Fig. 5. This observation indicates that our `AERIA` mechanism realizes the high usage efficiency for edge inference resources as performance benchmarks `Edgent`, `AMR`$^2$ and `Fixed Profit Rate = 1.0`. The `IAO` algorithm nonetheless performs the worst in terms of fulfilled DNN inference requests and edge inference resource utilization, because many edge inference requests are declined due to budget inaffordability.

## VII. Conclusion

This article introduces a market-oriented pricing strategy for edge inference services, built upon an auction mechanism administered by the AI service providers. From a practical perspective of Edge-AI market, we thoroughly address the personalized inference requirements of AI users, and the revenue incentives for the AI service provider. Specifically, we put forward an <u>A</u>uction-based <u>E</u>dge Inf<u>e</u>rence Pr<u>i</u>cing Mech<u>a</u>nism called `AERIA` for revenue maximization, which optimizes the DNN model partition, as well as edge inference pricing and resource allocation for revenue maximization. The proposed mechanism is theoretically proved to achieve several desirable properties, including competitiveness in revenue maximization, incentive compatibility, and envy-freeness. The superiority of `AERIA` over state-of-the-art approaches is demonstrated through extensive simulation experiments based on real-world datasets. Future work will extend `AERIA` by integrating recent advances in multi-device co-inference [48][49], enabling multiple end devices to collaborate on local inference and thereby improve the accuracy and efficiency of global inference tasks. To summarize, this research provides a significant contribution to the study of Edge-AI market.

## References

[1] C.-C. Lin, K.-Y. Chen, and L.-T. Hsieh, "Real-time charging scheduling of automated guided vehicles in cyber-physical smart factories using

feature-based reinforcement learning," *IEEE Transactions on Intelligent Transportation Systems*, vol. 24, no. 4, pp. 4016–4026, 2023.

[2] Q. He, Z. Feng, H. Fang, X. Wang, L. Zhao, Y. Yao, and K. Yu, "A blockchain-based scheme for secure data offloading in healthcare with deep reinforcement learning," *IEEE/ACM Transactions on Networking*, vol. 32, no. 1, pp. 65–80, 2024.

[3] N. R. Zema, E. Natalizio, L. Di Puglia Pugliese, and F. Guerriero, "3D trajectory optimization for multimission UAVs in smart city scenarios," *IEEE Transactions on Mobile Computing*, vol. 23, no. 1, pp. 1–11, 2024.

[4] C. Frenkel, D. Bol, and G. Indiveri, "Bottom-up and top-down approaches for the design of neuromorphic processing systems: Tradeoffs and synergies between natural and artificial intelligence," *Proceedings of the IEEE*, vol. 111, no. 6, pp. 623–652, 2023.

[5] H. Li, K. Ota, and M. Dong, "Learning IoT in edge: Deep learning for the Internet of Things with edge computing," *IEEE Network*, vol. 32, no. 1, pp. 96–101, 2018.

[6] X. Chen, J. Zhang, B. Lin, Z. Chen, K. Wolter, and G. Min, "Energy-efficient offloading for DNN-based smart IoT systems in cloud-edge environments," *IEEE Transactions on Parallel and Distributed Systems*, vol. 33, no. 3, pp. 683–697, 2022.

[7] Z. Xu, L. Zhao, W. Liang, O. F. Rana, P. Zhou, Q. Xia, W. Xu, and G. Wu, "Energy-aware inference offloading for DNN-driven applications in mobile edge clouds," *IEEE Transactions on Parallel and Distributed Systems*, vol. 32, no. 4, pp. 799–814, 2021.

[8] X. Tang, X. Chen, L. Zeng, S. Yu, and L. Chen, "Joint multiuser DNN partitioning and computational resource allocation for collaborative edge intelligence," *IEEE Internet of Things Journal*, vol. 8, no. 12, pp. 9511–9522, 2021.

[9] J. Li, W. Liang, Y. Li, Z. Xu, X. Jia, and S. Guo, "Throughput maximization of delay-aware DNN inference in edge computing by exploring DNN model partitioning and inference parallelism," *IEEE Transactions on Mobile Computing*, vol. 22, no. 5, pp. 3017–3030, 2023.

[10] Z. Zhuang, J. Chen, W. Xu, Q. Qi, S. Guo, J. Wang, L. Lu, H. Yang, and J. Liao, "DECC: Delay-aware edge-cloud collaboration for accelerating DNN inference," *IEEE Transactions on Emerging Topics in Computing*, vol. 13, no. 2, pp. 438–450, 2025.

[11] E. Li, L. Zeng, Z. Zhou, and X. Chen, "Edge AI: On-demand accelerating deep neural network inference via edge computing," *IEEE Transactions on Wireless Communications*, vol. 19, no. 1, pp. 447–457, 2020.

[12] S. Henna and A. Davy, "Distributed and collaborative high-speed inference deep learning for mobile edge with topological dependencies," *IEEE Transactions on Cloud Computing*, vol. 10, no. 2, pp. 821–834, 2022.

[13] F. Dong, H. Wang, D. Shen, Z. Huang, Q. He, J. Zhang, L. Wen, and T. Zhang, "Multi-exit DNN inference acceleration based on multi-dimensional optimization for edge intelligence," *IEEE Transactions on Mobile Computing*, vol. 22, no. 9, pp. 5389–5405, 2023.

[14] Z. Liu, J. Song, C. Qiu, X. Wang, X. Chen, Q. He, and H. Sheng, "Hastening stream offloading of inference via multi-exit DNNs in mobile edge computing," *IEEE Transactions on Mobile Computing*, vol. 23, no. 1, pp. 535–548, 2024.

[15] Z. Huang, F. Dong, D. Shen, J. Zhang, H. Wang, G. Cai, and Q. He, "Enabling low latency edge intelligence based on multi-exit DNNs in the wild," in *IEEE International Conference on Distributed Computing Systems (ICDCS)*, 2021, pp. 729–739.

[16] S. Teerapittayanon, B. McDanel, and H. Kung, "BranchyNet: Fast inference via early exiting from deep neural networks," in *International Conference on Pattern Recognition (ICPR)*, 2016, pp. 2464–2469.

[17] X. Wang, J. Ye, and J. C. Lui, "Decentralized scheduling and dynamic pricing for edge computing: A mean field game approach," *IEEE/ACM Transactions on Networking*, vol. 31, no. 3, pp. 965–978, 2023.

[18] L. Ma, X. Wang, X. Wang, L. Wang, Y. Shi, and M. Huang, "TCDA: Truthful combinatorial double auctions for mobile edge computing in Industrial Internet of Things," *IEEE Transactions on Mobile Computing*, vol. 21, no. 11, pp. 4125–4138, 2022.

[19] X. Wang, J. Ye, and J. C. S. Lui, "Mean field graph based D2D collaboration and offloading pricing in mobile edge computing," *IEEE/ACM Transactions on Networking*, vol. 32, no. 1, pp. 491–505, 2024.

[20] Z. Tang, F. Mou, J. Lou, W. Jia, Y. Wu, and W. Zhao, "Joint resource overbooking and container scheduling in edge computing," *IEEE Transactions on Mobile Computing*, vol. 23, no. 12, pp. 10 903–10 917, 2024.

[21] F. Ttncolu, A. Ben-Ameur, G. Dn, A. Araldo, and T. Chahed, "Dynamic time-of-use pricing for serverless edge computing with generalized hidden parameter markov decision processes," in *IEEE International Conference on Distributed Computing Systems (ICDCS)*, 2024, pp. 668–679.

[22] N. He, S. Yang, F. Li, L. Zhu, L. Sun, X. Chen, and X. Fu, "Incentive mechanism for resource trading in video analytic services using reinforcement learning," *IEEE Transactions on Services Computing*, vol. 17, no. 6, pp. 3803–3816, 2024.

[23] Q. Li, H. Yao, T. Mai, C. Jiang, and Y. Zhang, "Reinforcement-learning- and belief-learning-based double auction mechanism for edge computing resource allocation," *IEEE Internet of Things Journal*, vol. 7, no. 7, pp. 5976–5985, 2020.

[24] M. Liwang and X. Wang, "Overbooking-empowered computing resource provisioning in cloud-aided mobile edge networks," *IEEE/ACM Transactions on Networking*, vol. 30, no. 5, pp. 2289–2303, 2022.

[25] S. Li, J. Huang, and B. Cheng, "Resource pricing and demand allocation for revenue maximization in IaaS clouds: A market-oriented approach," *IEEE Transactions on Network and Service Management*, vol. 18, no. 3, pp. 3460–3475, 2021.

[26] K. He, X. Zhang, S. Ren, and J. Sun, "Deep residual learning for image recognition," in *IEEE Conference on Computer Vision and Pattern Recognition (CVPR)*, 2016, pp. 770–778.

[27] C. Szegedy, V. Vanhoucke, S. Ioffe, J. Shlens, and Z. Wojna, "Rethinking the inception architecture for computer vision," in *IEEE Conference on Computer Vision and Pattern Recognition (CVPR)*, 2016, pp. 2818–2826.

[28] S. Teerapittayanon, B. McDanel, and H. Kung, "Distributed deep neural networks over the cloud, the edge and end devices," in *IEEE International Conference on Distributed Computing Systems (ICDCS)*, 2017, pp. 328–339.

[29] W. Liao, J. Mao, C. Tan, and M. Zheng, "Deep learning-based dispatching conflict avoider in MEC systems," in *IEEE Wireless Communications and Networking Conference (WCNC)*, 2025, pp. 1–7.

[30] Y. Li, X. Zhang, T. Zeng, J. Duan, C. Wu, D. Wu, and X. Chen, "Task placement and resource allocation for edge machine learning: A GNN-based multi-agent reinforcement learning paradigm," *IEEE Transactions on Parallel and Distributed Systems*, vol. 34, no. 12, pp. 3073–3089, 2023.

[31] X. Huang, T. Han, and N. Ansari, "Smart grid enabled mobile networks: Jointly optimizing BS operation and power distribution," *IEEE/ACM Transactions on Networking*, vol. 25, no. 3, pp. 1832–1845, 2017.

[32] Y. Li, H. C. Ng, L. Zhang, and B. Li, "Online cooperative resource allocation at the edge: A privacy-preserving approach," in *IEEE International Conference on Network Protocols (ICNP)*, 2020, pp. 1–11.

[33] S. Hou, W. Ni, S. Zhao, B. Cheng, S. Chen, and J. Chen, "Frequency-reconfigurable cloud versus fog computing: An energy-efficiency aspect," *IEEE Transactions on Green Communications and Networking*, vol. 4, no. 1, pp. 221–235, 2020.

[34] A. V. Goldberg, J. D. Hartline, A. R. Karlin, M. Saks, and A. Wright, "Competitive auctions," *Games and Economic Behavior*, vol. 55, no. 2, pp. 242–269, 2006.

[35] A. V. Goldberg and J. D. Hartline, "Competitiveness via consensus," in *Annual ACM-SIAM Symposium on Discrete Algorithms (SODA)*, 2003, pp. 215–222.

[36] H. Moulin and S. Shenker, "Strategyproof sharing of submodular costs: Budget balance versus efficiency," *Economic Theory*, vol. 18, no. 3, pp. 511–533, 2001.

[37] M. Lewis, Y. Liu, N. Goyal, M. Ghazvininejad, A. Mohamed, O. Levy, V. Stoyanov, and L. Zettlemoyer, "BART: Denoising sequence-to-sequence pre-training for natural language generation, translation, and comprehension," in *Annual Meeting of the Association for Computational Linguistics (ACL)*, 2020, pp. 7871–7880.

[38] D. Elliott, S. Frank, K. Sima'an, and L. Specia, "Multi30K: Multilingual English-German image descriptions," in *ACL Workshop on Vision and Language (VL)*, 2016, pp. 70–74.

[39] A. Krizhevsky, I. Sutskever, and G. E. Hinton, "ImageNet classification with deep convolutional neural networks," *Communications of the ACM*, vol. 60, no. 6, pp. 84–90, 2017.

[40] K. Simonyan and A. Zisserman, "Very deep convolutional networks for large-scale image recognition." in *International Conference on Learning Representations (ICLR)*, 2015, pp. 1–14.

[41] A. Krizhevsky, G. Hinton *et al.*, "Learning multiple layers of features from tiny images," 2009.

[42] "Pricing calculator - configure and estimate the costs for Azure products," https://azure.microsoft.com/en-us/pricing/calculator/.

[43] F. Liu, Z. Zhou, H. Jin, B. Li, B. Li, and H. Jiang, “On arbitrating the power-performance tradeoff in SaaS clouds,” *IEEE Transactions on Parallel and Distributed Systems*, vol. 25, no. 10, pp. 2648–2658, 2014.

[44] A. Fresa and J. P. Champati, “Offloading algorithms for maximizing inference accuracy on edge device in an edge intelligence system,” *IEEE Transactions on Parallel and Distributed Systems*, vol. 34, no. 7, pp. 2025–2039, 2023.

[45] X. Zhang, N. Rajabi, K. Duh, and P. Koehn, “Machine translation with large language models: Prompting, few-shot learning, and fine-tuning with QLoRA,” in *ACL Conference on Machine Translation (WMT)*, 2023, pp. 468–481.

[46] J. Yuan, C. Yang, D. Cai, S. Wang, X. Yuan, Z. Zhang, X. Li, D. Zhang, H. Mei, X. Jia, S. Wang, and M. Xu, “Mobile foundation model as firmware: The way towards a unified mobile AI landscape,” in *ACM Annual International Conference on Mobile Computing and Networking (MobiCom)*, 2024, pp. 1–17.

[47] Y. Kaya, S. Hong, and T. Dumitras, “Shallow-deep networks: Understanding and mitigating network overthinking,” in *International Conference on Machine Learning (ICML)*, vol. 97, 2019, pp. 3301–3310.

[48] D. Wen, P. Liu, G. Zhu, Y. Shi, J. Xu, Y. C. Eldar, and S. Cui, “Task-oriented sensing, computation, and communication integration for multi-device edge AI,” *IEEE Transactions on Wireless Communications*, vol. 23, no. 3, pp. 2486–2502, 2024.

[49] D. Wen, X. Jiao, P. Liu, G. Zhu, Y. Shi, and K. Huang, “Task-oriented over-the-air computation for multi-device edge AI,” *IEEE Transactions on Wireless Communications*, vol. 23, no. 3, pp. 2039–2053, 2024.

# *Supplementary Material for* "Dynamic Pricing for On-Demand DNN Inference in the Edge-AI Market"

Songyuan Li, Jia Hu, Geyong Min Haojun Huang, and Jiwei Huang

## Appendix A
## Proof of Proposition 1

*Proof.* To rigorously justify the Proposition 1, we break the proof into the following two cases.

- **Case 1:** *The bidding AI user* $u_i$ *whose* $\widetilde{a}_{i,t} = \arg\max\left(\widetilde{a}_{1,t}, \ldots, \widetilde{a}_{N(t),t}\right)$ *is not admitted to edge inference acceleration by* ROSO *auction.*
  - ◇ Since none of revenue is earned from the bidding AI user $u_i$ in the ROSO auction, thus the upper bound of revenue $\widetilde{\Pi}(\mathcal{U}_t)$ by ROSO remains unchanged with no reduction.
- **Case 2:** *The bidding AI user* $u_i$ *whose* $\widetilde{a}_{i,t} = \arg\max\left(\widetilde{a}_{1,t}, \ldots, \widetilde{a}_{N(t),t}\right)$ *is admitted to edge inference acceleration by* ROSO *auction.*
  - ◇ In the worst-case scenario, the AI service provider sells $\widetilde{a}_{i,t}$ fewer edge-inference FLOPS units due to the absence of the bidding AI user $u_i$. Correspondingly, the upper bound of revenue $\widetilde{\Pi}(\mathcal{U}_t)$ by ROSO auction is reduced at most a factor of $\left(\widetilde{\Phi}_t - \zeta\right)/\widetilde{\Phi}_t$.

To put the above two cases into a nutshell, we complete the proof. □

## Appendix B
## Proof of Lemma 1

*Proof.* Fig. 1 demonstrates the probability distribution (*y-axis*) and value range (*upper x-axis*) for the exponent fraction of randomized consensus estimate functions $f\left(\Pi\left(\mathcal{U}_t\right)\right)$ and $f\left(\Pi\left(\mathcal{U}_t^{-i}\right)\right)$, with respect of different values $\varepsilon$. The exponent fractions $\log_y f\left(\Pi\left(\mathcal{U}_t\right)\right)$ and $\log_y f\left(\Pi\left(\mathcal{U}_t^{-i}\right)\right)$ are uniformly distributed over $\varepsilon \in [0,1]$, as indicated by the shadow area in Fig. 1. Meanwhile, the exponent fraction $\log_y f\left(\Pi\left(\mathcal{U}_t\right)\right)$ is continuously ranged from $\log_y \Pi\left(\mathcal{U}_t\right) - 1$ to $\log_y \Pi\left(\mathcal{U}_t\right)$. Similarly, the continuous value range of $\log_y f\left(\Pi\left(\mathcal{U}_t^{-i}\right)\right)$ is between $\left(\log_y \Pi\left(\mathcal{U}_t^{-i}\right) - 1, \log_y \Pi\left(\mathcal{U}_t^{-i}\right)\right]$.

Remarkably, the first subfigure (*Case 2a*) and the third subfigure (*Case 2b*) constructively characterize the probability distribution and value range for $\log_y f\left(\Pi\left(\mathcal{U}_t^{-i}\right)\right)$ in two cases. Therefore, it follows that:

Songyuan Li, Jia Hu, and Geyong Min are with the Department of Computer Science, Faculty of Environment, Science and Economy, University of Exeter, Exeter EX4 4PY, U.K. (e-mail:{S.Y.Li, J.Hu, G.Min}@exeter.ac.uk).

Haojun Huang is with the School of Electronic Information and Communications, Huazhong University of Science and Technology, Wuhan 430074, China (e-mail: hjhuang@hust.edu.cn).

Jiwei Huang is with the Beijing Key Laboratory of Petroleum Data Mining, China University of Petroleum, Beijing 102249, China (e-mail: huangjw@cup.edu.cn).

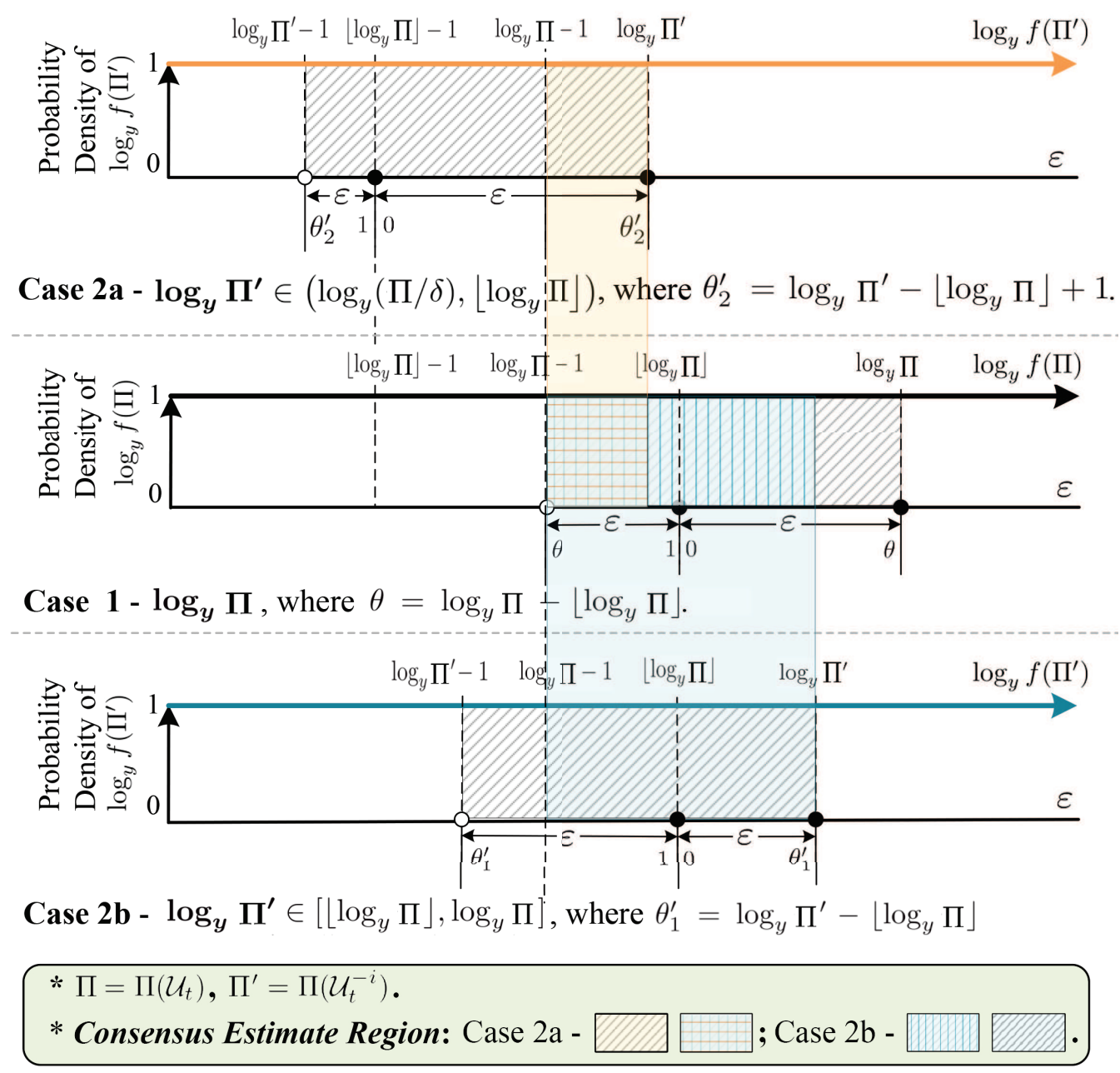

Fig. A.1: Probability Distribution and Value Range of Randomized Functions $\log_y f\left(\Pi\left(\mathcal{U}_t\right)\right)$ and $\log_y f\left(\Pi\left(\mathcal{U}_t^{-i}\right)\right)$.

- **Case 1:** $\log_y \Pi\left(\mathcal{U}_t\right)$, $\theta = \log_y \Pi - \lfloor\log_y \Pi\rfloor$.

$$\log_y f\left(\Pi\left(\mathcal{U}_t\right)\right) = \lfloor\log_y \Pi\left(\mathcal{U}_t\right)\rfloor + \varepsilon, \forall\varepsilon \in [0,\theta],$$
$$\log_y f\left(\Pi\left(\mathcal{U}_t\right)\right) = \lfloor\log_y \Pi\left(\mathcal{U}_t\right)\rfloor + \varepsilon - 1, \forall\varepsilon \in (\theta,1].$$
$$\Longrightarrow \log_y f\left(\Pi\left(\mathcal{U}_t\right)\right)|_{\varepsilon=\theta} = \log_y \Pi\left(\mathcal{U}_t\right),$$
$$\log_y f\left(\Pi\left(\mathcal{U}_t\right)\right)|_{\varepsilon=0,1} = \lfloor\log_y \Pi\left(\mathcal{U}_t\right)\rfloor,$$
$$\lim_{\varepsilon\to\theta^+} \log_y f\left(\Pi\left(\mathcal{U}_t\right)\right) = \log_y \Pi\left(\mathcal{U}_t\right) - 1.$$
$$\Longrightarrow \log_y f\left(\Pi\left(\mathcal{U}_t\right)\right) \in \left(\log_y \Pi\left(\mathcal{U}_t\right) - 1, \log_y \Pi\left(\mathcal{U}_t\right)\right].$$

- **Case 2a:** $\log_y \Pi\left(\mathcal{U}_t^{-i}\right) \in \left[\log_y(\Pi\left(\mathcal{U}_t\right)/\delta), \lfloor\log_y \Pi\left(\mathcal{U}_t\right)\rfloor\right)$, $\theta_2' = \log_y \Pi' - \lfloor\log_y \Pi\rfloor + 1$.

$$\log_y f\left(\Pi\left(\mathcal{U}_t^{-i}\right)\right) = \lfloor\log_y \Pi\left(\mathcal{U}_t\right)\rfloor + \varepsilon - 1, \forall\varepsilon \in [0,\theta_2'],$$
$$\log_y f\left(\Pi\left(\mathcal{U}_t^{-i}\right)\right) = \lfloor\log_y \Pi\left(\mathcal{U}_t\right)\rfloor + \varepsilon - 2, \forall\varepsilon \in (\theta_2',1].$$
$$\Longrightarrow \log_y f\left(\Pi\left(\mathcal{U}_t^{-i}\right)\right)|_{\varepsilon=\theta_2'} = \log_y \Pi\left(\mathcal{U}_t^{-i}\right),$$
$$\log_y f\left(\Pi\left(\mathcal{U}_t^{-i}\right)\right)|_{\varepsilon=0,1} = \lfloor\log_y \Pi\left(\mathcal{U}_t\right)\rfloor - 1,$$
$$\lim_{\varepsilon\to\theta_2'^+} \log_y f\left(\Pi\left(\mathcal{U}_t^{-i}\right)\right) = \log_y \Pi\left(\mathcal{U}_t^{-i}\right) - 1.$$
$$\Longrightarrow \log_y f\left(\Pi\left(\mathcal{U}_t^{-i}\right)\right) \in \left(\log_y \Pi\left(\mathcal{U}_t^{-i}\right) - 1, \log_y \Pi\left(\mathcal{U}_t^{-i}\right)\right].$$

- **Case 2b:** $\log_y \Pi\left(\mathcal{U}_t^{-i}\right) \in \left[\lfloor\log_y \Pi\left(\mathcal{U}_t\right)\rfloor, \log_y \Pi\left(\mathcal{U}_t\right)\right]$, $\theta_1' = \log_y \Pi' - \lfloor\log_y \Pi\rfloor$.

$$\log_y f\left(\Pi\left(\mathcal{U}_t^{-i}\right)\right) = \lfloor\log_y \Pi\left(\mathcal{U}_t\right)\rfloor + \varepsilon, \forall\varepsilon \in [0,\theta_1'],$$
$$\log_y f\left(\Pi\left(\mathcal{U}_t^{-i}\right)\right) = \lfloor\log_y \Pi\left(\mathcal{U}_t\right)\rfloor + \varepsilon - 1, \forall\varepsilon \in (\theta_1',1].$$

$$\Longrightarrow \log_y f\left(\Pi\left(\mathcal{U}_t^{-i}\right)\right)|_{\varepsilon=\theta_1'} = \log_y \Pi\left(\mathcal{U}_t^{-i}\right),$$
$$\log_y f\left(\Pi\left(\mathcal{U}_t^{-i}\right)\right)|_{\varepsilon=0,1} = \left\lfloor \log_y \Pi\left(\mathcal{U}_t\right)\right\rfloor,$$
$$\lim_{\varepsilon\to\theta_1'^+} \log_y f\left(\Pi\left(\mathcal{U}_t^{-i}\right)\right) = \log_y \Pi\left(\mathcal{U}_t^{-i}\right) - 1.$$
$$\Longrightarrow \log_y f\left(\Pi\left(\mathcal{U}_t^{-i}\right)\right) \in \left(\log_y \Pi\left(\mathcal{U}_t^{-i}\right) - 1, \log_y \Pi\left(\mathcal{U}_t^{-i}\right)\right].$$

By Definition 2, a consensus estimate is reached for the AI user $u_i \Leftrightarrow f\left(\Pi\left(\mathcal{U}_t\right)\right) = f\left(\Pi\left(\mathcal{U}_t^{-i}\right)\right)$, $\forall \Pi\left(\mathcal{U}_t^{-i}\right) \in \left[\Pi\left(\mathcal{U}_t\right)/\delta, \Pi\left(\mathcal{U}_t\right)\right]$. Correspondingly, it can be observed from Fig. 1 that the *overlap shadow area* in subfigures represents such *a consensus estimate region*, which fulfills $f\left(\Pi\left(\mathcal{U}_t\right)\right) = f\left(\Pi\left(\mathcal{U}_t^{-i}\right)\right)$ in *Case 2a* and *Case 2b*. Specifically speaking, $f\left(\Pi\left(\mathcal{U}_t\right)\right) = f\left(\Pi\left(\mathcal{U}_t^{-i}\right)\right) \le \Pi\left(\mathcal{U}_t^{-i}\right)$ holds true when $\varepsilon \in (\theta, \theta_2']$ in *Case 2a*, and $\varepsilon \in [0, \theta_1'] \cup (\theta, 1]$ in *Case 2b*. In other words,

- **Case 1** ($f\left(\Pi\left(\mathcal{U}_t\right)\right)$) ***v.s.*** **Case 2a** ($f\left(\Pi\left(\mathcal{U}_t^{-i}\right)\right)$):

$$f\left(\Pi\left(\mathcal{U}_t\right)\right) = f\left(\Pi\left(\mathcal{U}_t^{-i}\right)\right) = y^{\left\lfloor \log_y \Pi\left(\mathcal{U}_t\right)\right\rfloor + \varepsilon - 1},\ \forall \varepsilon \in (\theta, \theta_2'].$$
$\Longrightarrow$ *The consensus estimate is reached*, and
$$f\left(\Pi\left(\mathcal{U}_t\right)\right)|_{\varepsilon=\theta_2'} = f\left(\Pi\left(\mathcal{U}_t^{-i}\right)\right)|_{\varepsilon=\theta_2'} = \Pi\left(\mathcal{U}_t^{-i}\right),$$
$$\lim_{\varepsilon\to\theta^+} f\left(\Pi\left(\mathcal{U}_t\right)\right) = f\left(\Pi\left(\mathcal{U}_t^{-i}\right)\right)|_{\varepsilon=\theta} = \Pi\left(\mathcal{U}_t\right)/y.$$
$$\Longrightarrow f\left(\Pi\left(\mathcal{U}_t\right)\right) = f\left(\Pi\left(\mathcal{U}_t^{-i}\right)\right) \in \left(\Pi\left(\mathcal{U}_t\right)/y, \Pi\left(\mathcal{U}_t^{-i}\right)\right],\ \forall \varepsilon \in (\theta, \theta_2'].$$

- **Case 1** ($f\left(\Pi\left(\mathcal{U}_t\right)\right)$) ***v.s.*** **Case 2b** ($f\left(\Pi\left(\mathcal{U}_t^{-i}\right)\right)$):

$$f\left(\Pi\left(\mathcal{U}_t\right)\right) = f\left(\Pi\left(\mathcal{U}_t^{-i}\right)\right) = y^{\left\lfloor \log_y \Pi\left(\mathcal{U}_t\right)\right\rfloor + \varepsilon},\ \forall \varepsilon \in [0, \theta_1'],$$
$$f\left(\Pi\left(\mathcal{U}_t\right)\right) = f\left(\Pi\left(\mathcal{U}_t^{-i}\right)\right) = y^{\left\lfloor \log_y \Pi\left(\mathcal{U}_t\right)\right\rfloor + \varepsilon - 1},\ \forall \varepsilon \in (\theta, 1].$$
$\Longrightarrow$ *The consensus estimate is reached*, and
$$f\left(\Pi\left(\mathcal{U}_t\right)\right)|_{\varepsilon=\theta_1'} = f\left(\Pi\left(\mathcal{U}_t^{-i}\right)\right)|_{\varepsilon=\theta_1'} = \Pi\left(\mathcal{U}_t^{-i}\right),$$
$$\lim_{\varepsilon\to\theta^+} f\left(\Pi\left(\mathcal{U}_t\right)\right) = f\left(\Pi\left(\mathcal{U}_t^{-i}\right)\right)|_{\varepsilon=\theta} = \Pi\left(\mathcal{U}_t\right)/y.$$
$$\Longrightarrow f\left(\Pi\left(\mathcal{U}_t\right)\right) = f\left(\Pi\left(\mathcal{U}_t^{-i}\right)\right) \in \left(\Pi\left(\mathcal{U}_t\right)/y, \Pi\left(\mathcal{U}_t^{-i}\right)\right],\ \forall \varepsilon \in [0, \theta_1'] \cup (\theta, 1].$$

To summarize, we complete the proof by inducing that a consensus estimate is reached for the AI user $u_i \Leftrightarrow f\left(\Pi\left(\mathcal{U}_t\right)\right) \le \Pi\left(\mathcal{U}_t^{-i}\right)$. □

## APPENDIX C
## PROOF OF THEOREM 1

*Proof.* From Lemma 1, we can draw the conclusion that

$$\begin{gathered}\textit{The randomized function } f(\varpi) \textit{ is a consensus estimate}\\ \textit{for all bidding AI users } u_i \in \mathcal{U}_t\\ \Updownarrow\\ f(\Pi\left(\mathcal{U}_t\right)) \le \min_{u_i\in\mathcal{U}_t} \Pi\left(\mathcal{U}_t^{-i}\right).\end{gathered} \tag{A.1}$$

According to Proposition 1, $\Pi\left(\mathcal{U}_t\right)/\delta \le \Pi\left(\mathcal{U}_t^{-i}\right) \le \Pi\left(\mathcal{U}_t\right)$ holds, thereby inducing that

$$f\left(\pi\left(\mathcal{U}_t\right)\right) \leqslant \Pi\left(\mathcal{U}_t\right)/\delta \Rightarrow f\left(\Pi\left(\mathcal{U}_t\right)\right) \leqslant \min_{u_i\in\mathcal{U}_t} \Pi\left(\mathcal{U}_t^{-i}\right). \tag{A.2}$$

By combining Eq. (**??**) and Eq. (**??**), we complete the proof by deriving that

$$\begin{gathered}f\left(\Pi\left(\mathcal{U}_t\right)\right) \leqslant \Pi\left(\mathcal{U}_t\right)/\delta\\ \Downarrow\\ \textit{The randomized function } f(\varpi) \textit{ is a consensus estimate}\\ \textit{for all bidding AI users } u_i \in \mathcal{U}_t.\end{gathered}$$

□

## APPENDIX D
## PROOF OF LEMMA 2

*Proof.* Consider a random variable $X = \log_y f(\varpi)$, and let $\Delta = \log_y(\varpi/y)$. By Definition 3, the random variable $W$ represents $\log_y \varpi$ rounded down to the nearest $w+\varepsilon$ for integer $w$. Therefore,

$$\Pr[X \le x + \Delta] = \Pr\left[\varepsilon' \le x\right], \tag{A.3}$$

where $\varepsilon'$ is chosen uniformly between $[0, 1]$. It suggests that $X$ is uniformly distributed between $x$ and $x+1$. That is, the randomized function $f(\varpi)$ is distributed identically to $\varpi \cdot y^{\varepsilon'-1}$. □

## APPENDIX E
## PROOF OF THEOREM 2

*Proof.* According to Lemma 2, the randomized function $f(\varpi)$ is distributed identically to $\varpi \cdot y^{\varepsilon'-1}$, where $\mathbf{Pr}\left[f\left(\Pi\left(\mathcal{U}_t\right)\right) = F\right] = 1/(F \cdot \ln y)$. Therefore, we can complete the proof by the following derivation

$$\begin{aligned}\mathbf{E}\left[f\left(\Pi\left(\mathcal{U}_t\right)\right)\right] &= \int_{\Pi(\mathcal{U}_t)/y}^{\min\limits_{u_i\in\mathcal{U}_t}\Pi\left(\mathcal{U}_t^{-i}\right)} \left(F \cdot \frac{1}{F\cdot\ln y}\right) dF\\ &\ge \int_{\Pi(\mathcal{U}_t)/y}^{\Pi(\mathcal{U}_t)/\delta} \left(F \cdot \frac{1}{F\cdot\ln y}\right) dF\\ &= \frac{\Pi\left(\mathcal{U}_t\right)}{\ln y}\cdot\left(\frac{1}{\delta} - \frac{1}{y}\right).\end{aligned} \tag{A.4}$$

where the inequality holds as $\Pi\left(\mathcal{U}_t\right)/\delta \le \min\limits_{u_i\in\mathcal{U}_t} \Pi\left(\mathcal{U}_t^{-i}\right)$. □